\documentclass[lettersize,journal]{IEEEtran}
\usepackage{ragged2e}  
\usepackage{amsmath,amsfonts}
\usepackage{algorithm}
\usepackage{algorithmic}
\usepackage{array}
\usepackage{graphicx}

\usepackage{adjustbox}
\newcommand{\tsb}{\textsubscript}
\usepackage{textcomp}
\usepackage{stfloats}
\usepackage[table,xcdraw]{xcolor}
\usepackage{url}
\usepackage{multirow}
\usepackage{booktabs}
\usepackage{verbatim}
\usepackage{bbding}
\usepackage{stfloats}
\usepackage[accsupp]{axessibility}

\usepackage{todonotes}

\usepackage{amsmath}
\usepackage{amssymb}
\usepackage{graphicx}
\usepackage{subfigure}
\usepackage{amsthm}

\newtheorem{definition}{Definition}
\newtheorem{theorem}{Theorem}

\usepackage{tikz}
\definecolor{dashed_color}{HTML}{D3F131}
\definecolor{solid_color}{HTML}{FF6000}

\definecolor{wine}{HTML}{CB0000} 
\definecolor{dark_green}{HTML}{009901}

\ifCLASSOPTIONcompsoc
  \usepackage[nocompress]{cite}
\else
  \usepackage{cite}
\fi
\usepackage[pagebackref,breaklinks,colorlinks]{hyperref}
\usepackage{lettrine}

\title{Prompt Disentanglement via Language Guidance and Representation Alignment for Domain Generalization}

\author{
De Cheng*,~\IEEEmembership{}
Zhipeng Xu*,~\IEEEmembership{} 
Xinyang Jiang,~\IEEEmembership{} 
Dongsheng Li,~\IEEEmembership{} 
Nannan Wang,~\IEEEmembership{Senior Member,~IEEE,}
Xinbo Gao,~\IEEEmembership{Fellow,~IEEE}
\thanks{This work was supported in part by the New Generation Artificial Intelligence-National Science and Technology Major Project under Grant No.2025ZD0123601, in part by the National Natural Science Foundation of China under Grant 62576262, U22A2096, 62376206; in part by the Key R\&D Program of Shaanxi Province under 2024GX-YBXM135, in part by Scientific and Technological Innovation Teams in Shaanxi Province under 2025RS-CXTD-011, in part by the Shaanxi Province Core Technology Research and Development Project under 2024QY2-GJHX-11,in part by the Fundamental Research Funds for the Central Universities under QTZX25083, QTZX23042.

De Cheng, Zhipeng Xu, Nannan Wang, and Xinbo Gao are with the School of Telecommunications Engineering, the State Key Laboratory of Integrated Services Networks (ISN), Xidian University, Xi'an, China (email: dcheng@xidian.edu.cn, xu\_zhipeng@stu.xidian.edu.cn, nnwang@xidian.edu.cn, xbgao@mail.xidian.edu.cn).
Xinyang Jiang and Dongsheng Li is with Microsoft Research Asia, Shanghai, China (email: xinyangjiang@microsoft.com, dongsli@microsoft.com).

De Cheng and Zhipeng Xu contribute equally.

Corresponding Author: Nannan Wang (nnwang@xidian.edu.cn) and Xinyang Jiang (xinyangjiang@microsoft.com).
}
}

\begin{document}
\maketitle
\begin{abstract}
Domain Generalization (DG) seeks to develop models that perform well on unseen target domains by learning domain-invariant representations. Recent advances in pre-trained Visual Foundation Models (VFMs), such as CLIP, have shown strong potential for enhancing DG through prompt tuning. However, existing VFM-based prompt tuning methods often focus on task-specific adaptation rather than disentangling domain-invariant features, leaving cross-domain generalization insufficiently explored.
In this paper, we address this challenge by fully leveraging the controllable and flexible language prompt in VFMs. Observing that the text modality is inherently rich in semantics and easier to disentangle, we propose a novel framework termed Prompt Disentanglement via Language Guidance and Representation Alignment (PADG). 
PADG first employs a large language model (LLM) to disentangle textual prompts into domain-invariant and domain-specific components, which then guide the learning of domain-invariant visual representations. To complement the limitations of text-only guidance, we further introduce the Worst Explicit Representation Alignment (WERA) module, which enhances visual invariance by simulating bounded domain shifts through learnable stylization prompts and aligning representations between original and perturbed samples.
Extensive experiments on mainstream DG benchmarks, including PACS, VLCS, OfficeHome, DomainNet, and TerraInc, demonstrate that PADG consistently outperforms existing state-of-the-art methods, validating its effectiveness in robust domain-invariant representation learning.
\end{abstract}

\begin{IEEEkeywords}
Domain generalization, disentangled representation learning, Wasserstein Distributional Robust Learning
\end{IEEEkeywords}

\section{Introduction}
\label{sec:intro}
\IEEEPARstart{M}{ost} machine learning (ML) algorithms strongly rely on the assumption that training and testing data are independently and identically distributed (\emph{i.i.d.})~\cite{lew2023gradient, cha2022domain}, which is often violated in real-world scenarios due to \emph{domain shift}.
Acquiring data from all possible distributions to train an ML model is not only costly but often infeasible.
To overcome the domain shift problem, the Domain Generalization (DG) task has been introduced, aiming to learn a generalized model to perform well on any unseen target domains only using limited source domain data for training. 
DG has long been an essential topic in the ML research community and has attracted increasing interest.

Disentangled representation learning has always been the focus of ML for decades and is also the key to the success of domain generalization.
Fortunately, recent studies highlight the potential of large-scale pre-trained Vision Foundation Models (VFMs), such as CLIP, which is trained on vast (image, text) pairs to capture rich semantic information. 
Thus, these VFMs are capable of encoding the semantic content of visual descriptions, regardless of the image styles, which is in line with the goal of DG, \emph{i.e.,} learning consistent visual semantic representations across diverse domains. 

Recent advancements have demonstrated that prompt-based tuning methods refine the input to the text/visual encoder, enabling VFMs to better adapt to downstream tasks, even with limited training data.
However, directly applying prompt tuning in the context of DG remains a challenge. 
This is because existing prompt tuning methods tune the foundation model to generate task-specific features, whereas DG requires the model to generate domain-invariant features that work well across unseen domains. 
Therefore, to apply prompt tuning to DG effectively, it is crucial to design prompts that can guide the foundation model in disentangling invariant features across all domains from those specific to certain domains.

\begin{figure}[t]
\centering
\includegraphics[width=0.485\textwidth]{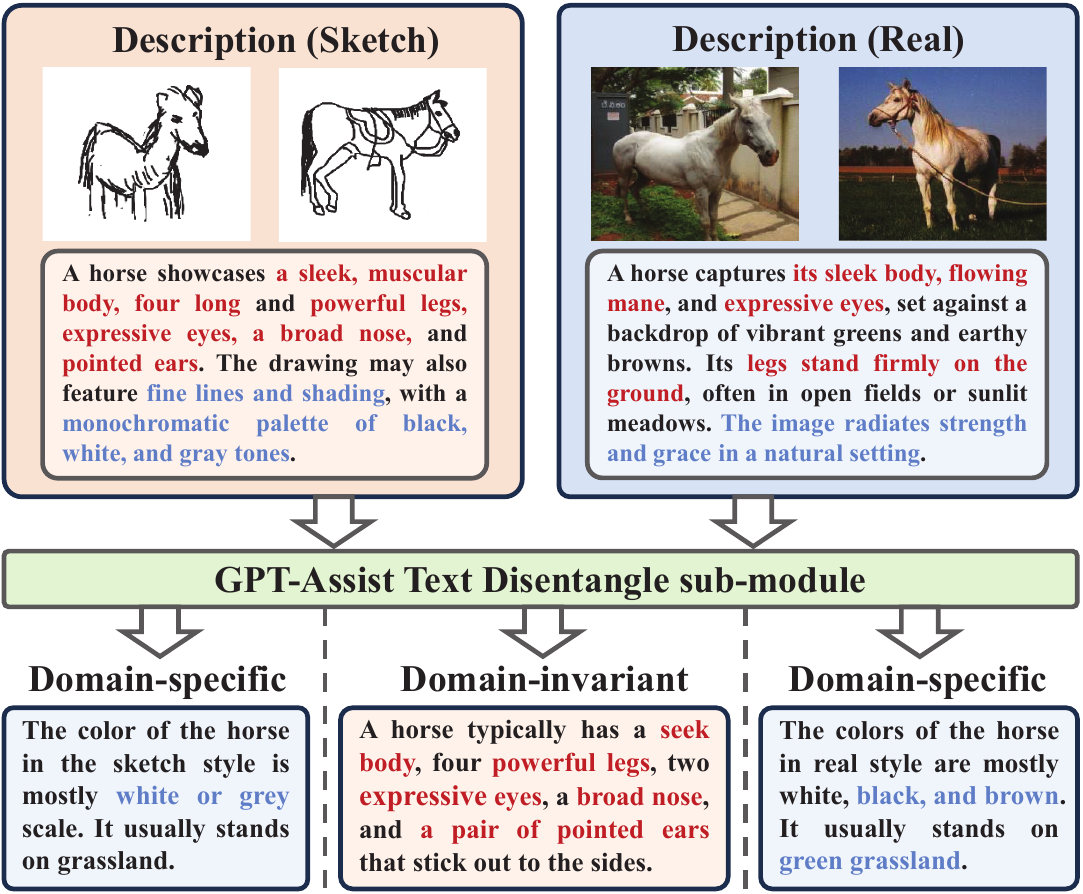}
\caption{
The upper panel of the figure highlights the contrasts between the sketch and real domains. 
These contrasts are subsequently disentangled into domain-invariant and domain-specific descriptions using the GPT-Assist text disentanglement sub-module, as shown in the lower panel of the figure.
}
\label{fig:motivation}
\end{figure}

In this paper, we address this challenge by fully leveraging a distinctive property of VFMs, namely their controllable and flexible language prompts. 
We argue that text prompts play a crucial role in guiding the disentanglement of visual features, since the text modality is inherently rich in semantic information and can clearly describe object-level concepts that are consistent across domains. 
For example, as illustrated in Fig.~\ref{fig:motivation}, a textual description of the class “horse” can be separated into domain-invariant details (e.g., the overall shape and body structure) and domain-specific details (e.g., its color or background). 
With the help of large language models (LLMs) such as GPT, such structured descriptions can be efficiently generated, providing reliable semantic cues that guide the learning of domain-invariant visual representations. 
However, while textual guidance effectively captures domain-invariant information, it is less capable of expressing domain-specific variations such as changes in viewpoint, or background context. 
Since these domain-specific aspects are difficult to describe precisely in language, the disentanglement guided only by text prompts tends to be incomplete, leaving residual domain information in the learned visual features and thereby reducing their domain invariance.
Motivated by Wasserstein distributionally robust learning, we further consider visual representation learning from a distributional perspective, where the source-domain data are conceptually mapped into a family of perturbed distributions. 
By applying controlled image transformations (e.g., color or geometric changes), we can simulate bounded domain shifts and enhance the invariance of the learned visual representations under these transformations.
This complementary strategy compensates for the limitations of text-only guidance and helps achieve more robust disentanglement between invariant and domain-specific factors across diverse domains.

As a result, we reframe prompt tuning from a disentanglement perspective and propose a unified framework, termed Prompt Disentanglement via Language Guidance and Representation Alignment for Domain Generalization (PADG). 
PADG integrates two complementary prompt-tuning-based disentanglement strategies for domain generalization. 
The first leverages text guidance by first disentangling textual prompts and then using them to guide the disentanglement of visual features, enabling the separation of domain-invariant and domain-specific components across modalities. 
The second complements this process by enhancing visual disentanglement through worst-case distributional alignment, which improves the robustness of learned visual representations against domain shifts.
Accordingly, PADG is implemented through a two-stage training framework that corresponds to these two strategies. 
In the first stage, the \textit{Cross-Modal Prompt Disentanglement (CMD)} module performs text-guided disentanglement to obtain domain-invariant and domain-specific representations across modalities. 
Specifically, domain-invariant and domain-specific textual descriptions for each category are generated using large language models (LLMs) and used for prompt tuning to learn disentangled textual embeddings, which then guide the visual encoder to learn semantically aligned visual features. 
The second stage employs the \textit{Worst Explicit Representation Alignment (WERA)} module to further enhance visual disentanglement through worst-case distributional alignment. 
Prompts reweight the mean and variance of each feature channel to produce stylized variants that mimic potential domain shifts, and an iterative optimization aligns visual representations between original and transformed samples, enabling the prompts to capture domain-invariant visual factors that remain stable under worst-case distributional perturbations. 
Through these two stages, PADG unifies cross-modal and visual disentanglement within a single prompt-tuning paradigm, thereby achieving robust domain generalization.

Effectively utilizing disentangled visual features for robust performance remains challenging. While most methods focus on domain-invariant features, we propose that domain-specific knowledge from similarly seen domains is equally vital for classifying unseen data. 
To bridge this gap, we introduce Domain-Specific Prototype Learning (DSPL), a novel ensemble strategy that combines domain-invariant and domain-specific knowledge for improved inference.
Specifically, DSPL constructs a prototype to collect domain-specific statistics for each class from each domain. 
During the inference stage, the domain-specific predictions are aggregated using as weights the similarity between a test data's instance-level visual feature and the source domain prototype.
The final output of the model is the weighted combination of the prediction of the domain-invariant visual representations and DSPL.

Our main contributions can be summarized as follows:

\begin{itemize} 

\item
We design a novel prompt-tuning framework for DG called Prompt Disentanglement via Language Guidance and Representation Alignment for Domain Generalization (PADG), which fully harnesses the potential of text modality to achieve disentanglement.

\item We propose the Worst Explicit Representation Alignment module, which adversarially stylizes visual features within a Wasserstein ball and enforces alignment between the original and stylized distributions, thereby enhancing the robustness and invariance of learned visual representations. Theoretical analysis in Sec.~\ref{sec:theoretical analysis} further provides a guarantee of generalization to unseen target domains.
 
\item We further devise a Domain-Specific Prototype Learning strategy, which integrates domain-specific statistics with domain-invariant predictions through a prototype-based ensemble mechanism to achieve more reliable inference.

\item Experiments conducted on mainstream DG datasets, including PACS, VLCS, OfficeHome, DomainNet, and TerraInc, demonstrate that the proposed PADG method outperforms the state-of-the-art DG methods.

\end{itemize}

This paper is an extension version of our conference paper~\cite{cheng2024disentangled} published in CVPR 2024.
To be specific, this version makes the following extensions: 
(1) We propose a novel prompt tuning module called Worst Explicit Representation Alignment (WERA), which adversarially stylizes training samples within a Wasserstein ball while preserving semantic fidelity through adaptive margin constraints, further enhancing the invariance of the visual representations.
(2) Theoretically, we guarantee our performance with respect to the unseen target domain via a brief analysis.
By analyzing the generalization bounds under well-defined assumptions, we demonstrate that the our proposed method maintains robust performance despite domain shifts.
(3) Compare to the conference version, besides multi-source domain generalization task, we further conduct additional experiments on single domain generalization.
Results demonstrate the substantial improvements and generalizability of our method compared to our previous work~\cite{cheng2024disentangled} and existing state-of-the-art methods.

\section{Related Work}~\label{Related Work}
\subsection{Domain Generalization.} 
Domain generalization focuses on developing models that can generalize to unseen target domains using only source domain data for training. 
Early research in this field has explored various methods for addressing domain shifts, including representation learning, meta-learning, data manipulation, and capturing causal relationships.
Representation learning seeks domain-invariant features by removing domain-specific information~\cite{bahng2020learning, li2018domain}.
For instance, \cite{wang2020crossdomain} uses disentangled representation learning to separate features into a domain-invariant content space and a domain-specific attribute space, enabling cross-domain generalization.
Meta-learning enhances generalization by dividing tasks into meta-train and meta-test sets, simulating domain shifts during training~\cite{zhang2021adaptive}.
Meanwhile, \cite{zhou2021domain,PAPT} shows that data manipulation, such as augmentations and data generation can significantly improve generalization when these variations effectively mimic real domain shifts.
Further research~\cite{arjovsky2020invariant, krueger2021outofdistribution} explores how capturing causal relationships can help address this challenge. 
Moreover, recent work~\cite{jiang2023domain, 9765363, 10.1145/3581783.3612073, he2024exploring} has also significantly contributed to this field.

With the rise of pre-trained Visual Foundation Models (VFMs), recent studies~\cite{IKI,StPR,RD-MLDG} have emphasized their strong generalization capabilities. 
Notably, \cite{gulrajani2020search} finds that simple ERM \cite{vapnik1999nature} outperforms the majority of early methods when applied to features from a pre-trained ResNet-50~\cite{he2015deep}. 
\cite{cha2022domain} proposes MIRO, which improves generalization by minimizing mutual information between pre-trained and oracle models. 
Meanwhile, \cite{niu2022domainunified} leverages CLIP~\cite{radford2021learning}, a large-scale vision-language pre-trained model, to extract domain-invariant representations through diverse prompt generation.
Additional contributions include gradient-based methods~\cite{lew2023gradient} and model-sample matching strategies~\cite{li2022simple}, which have further advanced domain generalization capabilities.
Building on these insights, our proposed method employs both CLIP and LLM to obtain the disentangled cross-modal features, facilitating robust model training.

\subsection{DG via Disentangled Representation Learning.}
The goal of Disentangled Representation Learning for DG is to decompose a feature representation into understandable compositions/sub-features, with one feature being the domain-invariant feature and the other domain-specific feature. 
Based on the choice of network structures and implementation mechanisms, disentanglement-based DG can mainly be categorized into three types: multi-component analysis, generative modeling, and causality-inspired methods.
Multi-component methods achieve disentanglement by learning separate domain-invariant and domain-specific features through dedicated network parameters~\cite{ding2017deep}.
From a generative perspective, recent work~\cite{zhang2022towards, nam2021reducing} shows that generative models can effectively learn disentangled representations by modeling the underlying data generation. 
Some works~\cite{zhang2015multi, gong2018causal} consider causality, as it can give information on how the system will behave under intervention.
Different from the existing disentanglement methods in DG, we propose a novel perspective to fully leverage the text modality to learn disentangled textual representations.
Then we achieve better disentanglement with the guidance of disentangled textual embeddings.

\subsection{DG via Wasserstein Distributionally Robust Learning.}
Distributionally Robust Optimization (DRO) formulates learning objectives that optimize performance under worst-case distributional shifts within a defined uncertainty set, providing theoretical guarantees against domain variations~\cite{sagawa2019distributionally}. 
DRO methods employ various discrepancy metrics to quantify distributional differences, including: (1) Wasserstein distance~\cite{sinha2017certifying, mehra2022certifying}, (2) $f$-divergences~\cite{namkoong2016stochastic, michel2021modeling}, and (3) maximum mean discrepancy~\cite{staib2019distributionally}.
However, despite these theoretical foundations, applying DRO to DG has yielded limited practical improvements~\cite{liu2021towards}.
The uncertainty set defined via the Wasserstein ball exhibits greater flexibility, which has led to extensive studies of Wasserstein Distributionally Robust Learning (WDRL)~\cite{sinha2017certifying, mohajerin2018data}.
Although WDRL provides a theoretically grounded alternative to empirical risk minimization for achieving robustness under distributional perturbations.
However, recent studies have questioned its practical effectiveness.
\cite{hu2018does} proved that when the WDRL is applied to classification tasks, the obtained classifier ends up being optimal for the observed training distribution, and the core of the proof lies in the over-flexibility of the built uncertainty set.
\cite{frogner2019incorporating} addressed this by using unlabeled data to constrain the distribution set.
Recent innovations include data geometry-based uncertainty sets \cite{liu2022distributionally} and topology-driven approaches \cite{qiao2023topology}. 
Our method tackles low-confidence predictions by exploring domain-specific uncertainty subsets and measuring distribution discrepancies using Wasserstein distance \cite{sinha2017certifying, mehra2022certifying}, enhancing robustness in Domain Generalization.

\begin{figure*}[!htbp]
\centering
\includegraphics[width=1.00\textwidth]{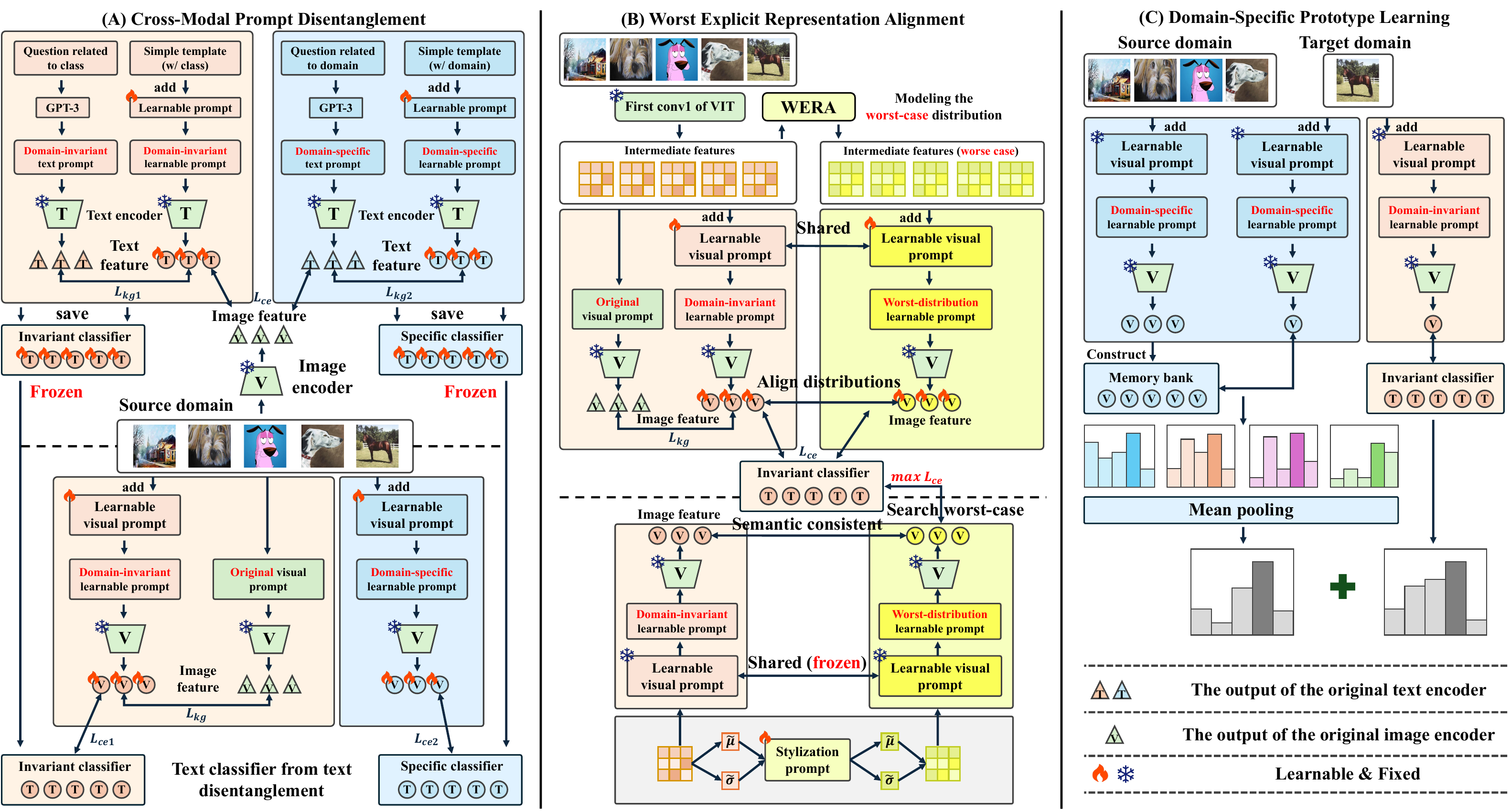}
\vspace{-4.0mm}
\caption{
\footnotesize{
Framework of the \textbf{PADG}. 
(A) \textbf{Cross-Modal Prompt Disentanglement (CMD)} consists of two sub-modules: 
\textbf{GPT-Assist Text Disentanglement (GAT)}, which generates domain-invariant and domain-specific textual embeddings, 
and \textbf{Image Disentanglement Guided by Text (IMT)}, which learns corresponding visual representations under textual guidance. 
(B) \textbf{Worst Explicit Representation Alignment (WERA)} transfers each training sample to a worst-case distribution by maximizing the empirical risk $L_{\mathrm{ce}}$, 
and aligns the visual representations between the original and stylized distributions to enhance cross-domain invariance. 
(C) \textbf{Domain-Specific Prototype Learning (DSPL)} employs a prototype-based ensemble strategy to exploit domain-specific knowledge during inference, where the final prediction is obtained by combining the domain-invariant prediction and the domain-specific prediction.
}}
\label{fig:example}
\vspace{-2.5mm}
\end{figure*}

\section{Methodology}
\noindent \textbf{Problem Definition}.  
Let $\mathcal{X}$ be the input (feature) space and $\mathcal{Y}$ the target (label) space. 
We define $\hat{\mathcal{D}}^S$ and $\hat{\mathcal{D}}^T$ as the sets corresponding to the source and target domains.
Specifically, $\hat{\mathcal{D}}^{S}=\{\hat{\mathcal{D}}^{S}_m\}_{m=1}^{N_s}$, where each $\hat{\mathcal{D}}^{S}_m$ denote a distribution over the input space $\mathcal{X}$, and $N_s$ is the total number of source domains. 
For each source domain, $\hat{\mathcal{D}}^S_m=\{(\mathbf{x}_i,y_i)\}_{i=1}^{N^s_m}$, where $(\mathbf{x}_i, y_i)$ denotes a data sample consisting of an input image $\mathbf{x}_i$ and its corresponding label $y_i$. 
The unseen target domains are similarly defined as $\hat{\mathcal{D}}^T=\{\hat{\mathcal{D}}^T_m\}_{m=1}^{N_t}$.
The objective of DG is to construct a model $\phi_{\theta}(\cdot)$ capable of achieving robust performance on target domains, utilizing only one or multiple source domain data during training.
The key challenge is to mitigate the domain discrepancies between the source and target data distribution. 
Note that, this study is specifically validated on the task of image classification.

\noindent \textbf{Overall Framework of our proposed PADG.} 
The framework of our PADG is shown in Fig.~\ref{fig:example}. 
PADG comprises two training stages, Cross-Modal Prompt Disentanglement (CMD) and Worst Explicit Representation Alignment (WERA), followed by an additional inference stage, Domain-Specific Prototype Learning (DSPL).
Specifically, as shown in Fig.~\ref{fig:example} (\textcolor{red}{A}), CMD is designed to achieve disentanglement in both the textual and visual modalities. 
It consists of two sub-modules: GPT-Assist Text Disentanglement (GAT), which learns domain-invariant and domain-specific textual embeddings, and Image Disentanglement Guided by Text (IMT), which leverages these textual embeddings to guide the separation of visual representations. 
In the second training stage, illustrated in Fig.~\ref{fig:example} (\textcolor{red}{B}), WERA generates diverse, challenging distributions by constructing worst-case samples under the current model and aligns visual representations between the original and transformed distributions, thereby enhancing the cross-domain invariance of the learned features. 
Finally, in the inference stage shown in Fig.~\ref{fig:example} (\textcolor{red}{C}), DSPL employs a prototype-based ensemble strategy that integrates domain-specific predictions with domain-invariant ones to produce the final output. 
Through this framework, PADG effectively captures and combines both domain-invariant and domain-specific knowledge, leading to more robust and generalized representations.

\subsection{Preliminaries}
\textbf{Prompt Tuning on CLIP Model.} 
We employ CLIP~\cite{radford2021learning} as the vision-language backbone for prompt tuning. 
It comprises a visual encoder $\mathbf{f}(\cdot)$ and a text encoder $\mathbf{g}(\cdot)$ that map images and text into a shared embedding space. 
During training, only the prompt are learnable, while both encoders remain frozen.

For a downstream task with $N_c$ categories, CLIP employs a fixed hand-craft prompt, denoted as $\mathbf{t}_k$ for the $k_{th}$ class, to generate its corresponding textual embedding $\mathbf{w}^{clip}_k=\mathbf{g}(\mathbf{t}_k)$, and the textual embeddings for all categories can be denoted as $\mathbf{W}^{clip}=\{\mathbf{w}^{clip}_k\}_{k=1}^{N_c}$. 
Given an image $\mathbf{x}_i$ with corresponding label $y_i$, the visual embedding can be obtained by the visual encoder $\mathbf{f}(\cdot)$ as: $\mathbf{z}^{clip}_i=\mathbf{f}(\mathbf{x}_i)$. 
After that, the prediction probability for image $\mathbf{x}_i$ is then computed as follows:
\begin{equation}
    p_k(y_i \mid \mathbf{z}^{clip}_i; \mathbf{W}^{clip})=\frac{\exp (<\mathbf{z}^{clip}_i,\mathbf{w}_{k}^{clip}> / \tau)}{\sum\limits_{k=1}^{N_{c}} \exp (<\mathbf{z}^{clip}_i, \mathbf{w}_{k}^{clip}> / \tau)},
    \label{ContrastLearning}
\end{equation}
where we can define $\mathbf{P}=[p_1, \cdots, p_k, \cdots, p_{N_c}]$ as the collection of probabilities for classifying $\mathbf{x}_i$, and $p_k$ in Eq.~\ref{ContrastLearning} represents the probability of $\mathbf{x}_i$ belonging to the $k_{th}$ category. 
$\tau$ is a hyper-parameter to control the sharpness of the output, and $<\cdot,\cdot>$ is the dot product which can be termed as the cosine similarity as the features are normalized.  
Combined with cross-entropy loss, we can fine-tune the CLIP model.

\noindent \textbf{Text Prompt Tuning}. 
Although Eq.~\ref{ContrastLearning} can be easily applied to Zero-Shot classification by employing a fixed hand-craft prompt, (\emph{i.e.}, $\mathbf{t}=$\emph{`a photo of a [class]'}), to generate the textual embedding, it can not be well adapted to the new downstream task. 
Therefore, the prompt tuning method is proposed by learning a set of continuous vectors for generating task-related textual embedding, \emph{e.g.}, CoOp~\cite{zhou2022learning}. Specifically, the learnable prompt for the input of the text encoder can be expressed as:
\begin{equation} \label{clip_function_2}
    \mathbf{t}_k=\left[\mathbf{v}_1, \mathbf{v}_2, \cdots, \mathbf{v}_L, \mathbf{CLS}_k\right], \quad \forall k \in 1 \sim N_c,
\end{equation}
where $\mathbf{v}_i\in \mathbb{R}^d$ is a learnable vector and $d$ is the prompt dimension, $L$ is the length of the prompt, and $\mathbf{CLS}_k$ is the class token for the $k_{th}$ text prompt $\mathbf{t}_k$. During training, only the learnable prompt is updated while the original visual and text encoder are frozen. The output of the text encoder can be represented as $\mathbf{w}_k=\mathbf{g}(\mathbf{t}_k)$. 
Specifically, we adopt the CoOp~\cite{zhou2022learning} mechanism to learn the text prompt enabling the CLIP model to better adapt to downstream tasks. 

\noindent \textbf{Visual Prompt Tuning}. 
To fine-tune the visual encoder $\mathbf{f}(\cdot)$, we adopt the deep VPT technique~\cite{jia2022visual} to insert a set of learnable prompts into the transformer layer of the visual encoder. 
VPT employs multiple layers of Transformers to capture the intricate relationship between images and text. Notably, it inserts learnable prompts between the patch embeddings and the class token at the input of each layer. 

The VPT model links the output of the last class token to a fully connected layer and applies a softmax function to generate a probability over classes. During training, the model employs cross-entropy loss to compare the predicted class probability with the ground-truth label. It only updates the learnable prompt parameters while keeping the pre-trained model parameters fixed.

\noindent \textbf{Wasserstein Distributionally Robust Learning}. 
To obtain a more invariant visual representation, we leverage the Wasserstein Distributionally Robust Learning technique (WDRL) by constructing a worst-case distribution set containing all distributions within a Wasserstein ball, centered on the distribution of source domains.
Then, by ensuring uniformly well performance across the distribution set, 
the learned visual representations can be invariant to more distributions, further mitigating the negative impacts of the domain shift.
We first provide the definition of Wasserstein distance in Eq.~\ref{function:wasserstein-distance}, and then based on this definition, we present the optimization objective of WDRL in Eq.~\ref{function:worst-case}.

\begin{definition}\label{definition:W1} (Optimal Transport Cost and Wasserstein-1 Distance~\cite{villani2009optimal, villani2021topics}).
Given two probability measures $\mathcal{D}^{S}$ and $\mathcal{D}^{\mathcal{T}}$ defined on a measurable space $\mathcal{X}$.
Let a class of measurable functions $\Phi=\{\phi:\|\phi\|_L \leq 1\}$ where
\begin{equation}
    \|\phi\|_{L}=\sup\limits_{x\neq x^{'}\in \mathcal{X}}\frac{|\phi(x)-\phi(x^{'})|}{c(x,x^{'})}
\end{equation}
is the Liptschitz semi-norm for real-valued continuous $\phi$ on $\mathcal{X}$ and some metric $c(\cdot,\cdot):\mathcal{X}\times \mathcal{X}\rightarrow \mathbb{R}_{+}$. In this case, the Kantorovich-Rubinstein theorem~\cite{Dudley_2002} yields the following result, with the Wasserstein distance $W_1$ defined as follows:
\begin{equation} 
\begin{aligned} \label{function:wasserstein-distance}
    W_1(\mathcal{D}^{S},\mathcal{D}^{\mathcal{T}})
    &=\sup\limits_{\|\phi\|_L\leq 1}|\int \phi d(\mathcal{D}^{S},\mathcal{D}^{\mathcal{T}})| \\
    &=\inf\limits_{s\in \Pi(\mathcal{D}^{S},\mathcal{D}^{\mathcal{T}})}\int_{\mathcal{X}\times \mathcal{X}}c(x,x^{'})d s(x,x^{'}),
\end{aligned}
\end{equation}
where $\Pi(\mathcal{D}^{S},\mathcal{D}^{\mathcal{T}})$ is a space of all joint probability measures on $\mathcal{X}\times \mathcal{X}$ with marginals $\mathcal{D}^{S}$ and $\mathcal{D}^{\mathcal{T}}$. 
\end{definition}


\begin{definition} (Wasserstein Distributionally Robust Learning~\cite{kuhn2019wasserstein, sinha2017certifying}).
Given $\mathcal{D}^S$ defined on a measurable space $\mathcal{X}$. Let $W_1(\cdot,\cdot)$ is the Wasserstein distance in Definition~\ref{definition:W1}. 
We consider the robustness region $\{\mathcal{D}^Q:W_1(\mathcal{D}^Q,\mathcal{D}^S)\leq 
\rho\}$, a $\rho$-neighborhood of the distribution $\mathcal{D}^S$ under the $W_1(\cdot,\cdot)$. The objective of the DRO~\cite{kuhn2019wasserstein} can be formulated as:
\begin{equation} \label{function:distribution_robust_learning}
    \inf\limits_{\theta\in \Theta}
    \left \{
    \sup\limits_{\mathcal{D}^{Q}}\{\mathbb{E}_{Q} 
    [\ell(\theta;\mathcal{D}^{Q})]\}
    \right \},
\end{equation}
where $\Theta$ is the model's parameter space and $\rho$ is the radius of Wasserstein-1 ball. $\ell$ is a bounded loss function: $\mathcal{Y}\times \mathcal{Y}\rightarrow \mathbb{R}_{+}$.
For deep networks and other complex models, the formulation of Eq.~\ref{function:distribution_robust_learning} is intractable with arbitrary $\rho$.
Instead, we consider its Lagrangian relaxation for a fixed penalty parameter $\gamma^{'} \geq 0$,
The optimization objective of WDRL~\cite{sinha2017certifying} can be defined as:
\begin{equation} \label{function:worst-case}
\inf\limits_{\theta\in \Theta} \left \{ \sup\limits_{\mathcal{D}^{Q}}\{\mathbb{E}_{Q} 
[\ell(\theta;\mathcal{D}^{Q})]-\gamma^{'} W_1(\mathcal{D}^Q,\mathcal{D}^S)\} \right \}.
\end{equation} 
Based on the~\cite{sinha2017certifying}, for large enough penalty $\gamma^{'}$ (by duality, small enough robustness $\rho$), the function in Eq.~\ref{function:worst-case} is strongly concave and hence easy to optimize if $\ell(\theta,\mathcal{D}^Q)$ is smooth.

\end{definition}

\begin{figure}[t]
\centering
\includegraphics[width=0.48\textwidth]{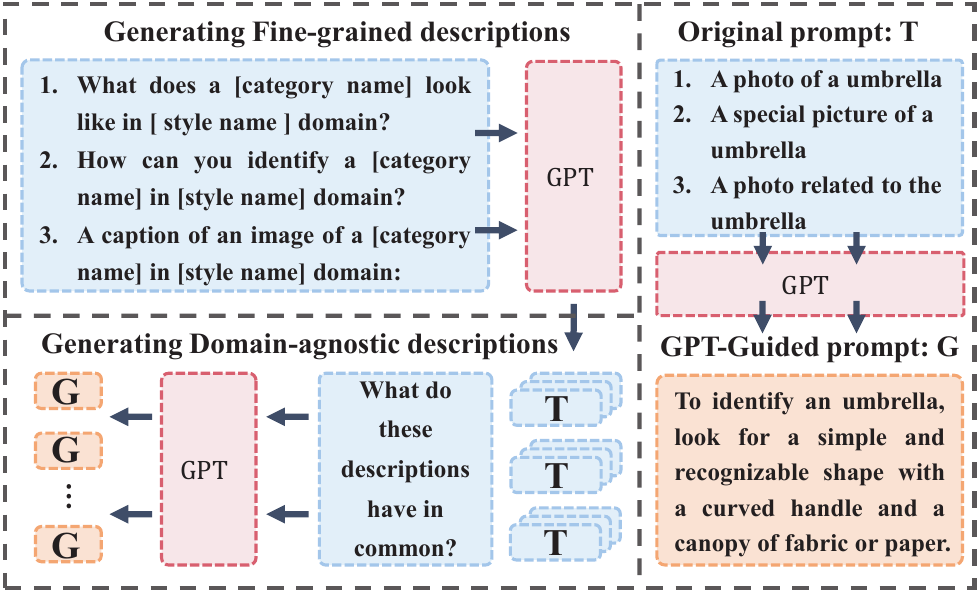}
\vspace{-3.0mm}
\caption{
The figure illustrates the two components of our GPT-Assist Text-Disentanglement sub-module. 
Firstly, fine-grained descriptions are generated for the several input questions. 
These descriptions are then summarized to produce domain-invariant descriptions. 
The right half of the figure contrasts the hand-craft descriptions with the domain-invariant descriptions.
}
\label{pic:disentanlement}
\vspace{-2.5mm}
\end{figure}

\subsection{Cross-Modal Prompt Disentanglement} \label{text_image_side}
To achieve disentanglement for both text and image modality, we propose a Cross-Modal Prompt Disentanglement module (CMD).
The text modality in the visual foundation model is inherently rich in semantic information and can be more easily disentangled.
Thereby we first introduce our GPT-Assist Text-Disentanglement sub-module (GAT) to obtain disentangled textual embeddings by leveraging the powerful LLM in Sec.~\ref{Text_side}.
Subsequently, with the guidance of disentangled textual embeddings, we introduce our Image-Disentanglement Guided by Text sub-module (IMT) in Sec.~\ref{Image_side}.
\subsubsection{GPT-Assist Text-Disentanglement} \label{Text_side}
To achieve the disentanglement of text modality, we first adopt GPT-3 to generate fine-grained descriptions for each class and each domain.  
Firstly, we design a unified series of templates as the language command for GPT-3 to generate fine-grained descriptions:
1) \emph{`What does a [class] look like in [domain]'}; 
2) \emph{`How can you identify an [class] in [domain]'}; 
3) \emph{`A caption of an image of an [class] in [domain]'}, as well as some other prompts: \emph{`Please provide a detailed description of the object under this [domain] and [class]'}. 
Subsequently, following the generation of an extensive description set, we utilize the GPT-3 once more to identify cross-domain attribute invariance through latent semantic analysis, finally obtaining more robust and domain-invariant descriptions for each class. 
Meanwhile, we also generate descriptions for each domain in the same manner.
Fig.~\ref{pic:disentanlement} shows the process of text disentanglement by GPT-3.
We can observe that the obtained domain-invariant description for the `umbrella' exhibits substantially richer semantic content than hand-craft descriptions.

After obtaining the domain-invariant description for each class, we generate the domain-invariant embedding for each class by the text encoder $\mathbf{g}(\cdot)$ of CLIP. Such domain-invariant embedding $\mathbf{w}_{k}^{clip}$ will later serve as guidance for learning domain-invariant text prompts $\mathbf{t}^I_k$. Specifically, we adopt the $L2$ distillation loss~\cite{hinton2015distilling} as follows:
\begin{equation}
\begin{aligned}
    \mathcal{L}^{T(I)}_{kg} = \frac{1}{N_c} &\sum_{k=1}^{N_c} ||\tilde{\mathbf{w}}_k-{\mathbf{w}}_{k}^{clip}||_2^2, \vspace{2.0mm} \\
    where \quad \tilde{\mathbf{w}}_k = \mathbf{g}(\mathbf{t}^I_k), \  &\mathbf{t}^I_k=[\mathbf{v}_1, \mathbf{v}_2, \cdots,\mathbf{v}_L, \mathbf{CLS}_k],
\end{aligned}
\end{equation}
$N_c$ is the number of class, $||\cdot||_2^2$ is the $L2$ norm. Note that ${\mathbf{w}}_{k}^{clip}$ is obtained by the original CLIP model. To learn the domain-invariant text prompt, we simultaneously optimize the above-mentioned distillation loss and the contrastive learning objective $\mathcal{L}^{T(I)}_{ce}$ between the $\tilde{\mathbf{w}}_k$ and visual embeddings $\mathbf{z}_i^{clip}$ from fixed visual encoder as follows:
\begin{equation}
    p^T_k(y_i \mid \mathbf{z}^{clip}_i; \tilde{\mathbf{W}})=\frac{\exp (<\mathbf{z}^{clip}_i,\tilde{\mathbf{w}}_{k}> / \tau)}{\sum\limits_{k=1}^{N_{c}} \exp (<\mathbf{z}^{clip}_i, \tilde{\mathbf{w}}_{k}> / \tau)},
    \label{ContrastLearning_CoOp}
\end{equation}
\begin{equation} \label{Lce1_represetation}
    \mathcal{L}^{T(I)}_{ce}=-\frac{1}{B} \sum_{i=1}^B \mathbf{y}_i \cdot \log \mathbf{P}^{T(I)}(y_i \mid \mathbf{z}^{clip}_i; \tilde{\mathbf{W}}), 
\end{equation}
where $\mathbf{y}_i \in \mathbb{R}^{N_c}$ in Eq.~\ref{Lce1_represetation} represents one-hot label and $N_c$ is the label dimension. 
For each input image $\mathbf{x}_i$, the actual label $\mathbf{y}_i$ is assigned as $[1, 0, \cdots, 0]$.
$\mathbf{P}^{T(I)} \in \mathbb{R}^{N_c}$ in Eq.~\ref{Lce1_represetation} is the collection of probabilities $p_k^{T}$ and $B$ denotes the batch size.

Similarly, we learn the domain-specific textual embedding $\tilde{\mathbf{h}}_m$ for $m_{th}$ domain with the learnable prompt $\mathbf{t}_m^S$ under the guidance of $\mathbf{h}_{m}^{clip}$ generated by domain-specific textual descriptions.
Additionally, we define the textual embedding for all domains as $\tilde{\mathbf{H}}=\{\tilde{\mathbf{h}}_m\}_{m=1}^{N_s}$.
In line with the training method for domain-invariant text prompts, we employ contrastive learning objective $\mathcal{L}^{T(S)}_{ce}$ and $L2$ distillation loss $\mathcal{L}^{T(S)}_{kg}$. 
Finally, the overall loss function can be formulated as:
\begin{equation} \label{loss_1} 
\mathcal{L}^{T}_{all}=\mathcal{L}^{T(I)}_{ce}+\mathcal{L}^{T(S)}_{ce}+\alpha_1*(\mathcal{L}^{T(I)}_{kg}+\mathcal{L}^{T(S)}_{kg}),
\end{equation}
where $\alpha_1$ is the hyper-parameter to balance the loss terms. 
In summary, the domain-invariant and domain-specific text prompts are optimized using cross-entropy losses $\mathcal{L}_{ce}^{T(I)}$ and $\mathcal{L}_{ce}^{T(S)}$, regularized by normalization terms $\mathcal{L}_{kg}^{T(I)}$ and $\mathcal{L}_{kg}^{T(S)}$, respectively.  
During the backward-propagation stage, the domain-invariant text prompt $\{\mathbf{t}^I_k\}_{k=1}^{N_c}$ and domain-specific text embedding $\{\mathbf{t}^S_m\}_{m=1}^{N_s}$ are jointly updated by the Eq.~\ref{loss_1}.

\subsubsection{Image-Disentanglement Guided by Text} \label{Image_side}
After achieving the text disentanglement, we keep the text encoder and its domain-invariant and domain-specific text prompt fixed. 
Then, we perform our Image-Disentanglement Guided by Text (IMT) sub-module under the guidance of the disentangled textual embeddings.
Specifically, we adopt the deep VPT as the visual encoder to extract the domain-invariant and domain-specific visual features, denoted as $\tilde{\mathbf{z}}^{I}_{i}$ and $\tilde{\mathbf{z}}^{S}_{i}$:
\begin{equation} \label{image_feature}
\left \{
\begin{array}{ll}
    \tilde{\mathbf{z}}^{I}_{i}=\mathbf{f}_I(\mathbf{x}_i; \tiny{\mathbf{E}_I}), 
    \vspace{2.0mm} \\
    \tilde{\mathbf{z}}^{S}_{i}=\mathbf{f}_s(\mathbf{x}_i; \tiny{\mathbf{E}_s}),
\end{array} \right.
\end{equation}
where $\mathbf{E}_I$ and $\mathbf{E}_s$ represent the learnable visual prompts in the domain-invariant and domain-specific visual encoders ($i.e.$, $\mathbf{f}_I(\cdot)$ and $\mathbf{f}_s(\cdot)$), respectively.
To finetune the visual encoder $\mathbf{f}_I(\cdot)$ in Eq.~\ref{image_feature}, we adopt the following loss terms: 1) the contrastive learning objective $\mathcal{L}^{V(I)}_{ce}$ between $\tilde{\mathbf{z}}_i^I$ and $\tilde{\mathbf{w}}_k$ to optimize $\mathbf{E}_I$; 2) the domain confusion regularization $\mathcal{L}_{mix}$:
\begin{equation}  
    p_m^V(y_d \mid \tilde{\mathbf{z}}^{I}_{i}; \tilde{\mathbf{H}})=\frac{\exp (<\tilde{\mathbf{z}}^{I}_{i}, \tilde{\mathbf{h}}_{m}> / \tau)}{\sum\limits_{m=1}^{N_s} \exp (<\tilde{\mathbf{z}}^{I}_{i}, \tilde{\mathbf{h}}_{m}> / \tau)},
    \label{DI-Prediction}   
\end{equation}
\begin{equation} \label{confusion_loss}
    \mathcal{L}_{mix}=-\frac{1}{B} \sum_{i=1}^{B} \mathbf{o}_i \cdot \log \mathbf{P}^{V(S)}(y_d \mid \tilde{\mathbf{z}}^{I}_{i}; \tilde{\mathbf{H}}),
\end{equation}
where $y_d$ in Eq.~\ref{DI-Prediction} and Eq.~\ref{confusion_loss} is the domain label for each input image $\mathbf{x}_i$. 
$\mathbf{P}^{V(S)} \in \mathbb{R}^{N_s}$ in Eq.~\ref{confusion_loss} is the collection of probabilities of $p_m^V$.
To train the domain confusion regulation, we set one-hot domain label $\mathbf{o}_i \in \mathbb{R}^{N_s}$ in Eq.~\ref{confusion_loss} to $[1/N_s, 1/N_s, \cdots, 1/N_s]$ for all the images from different domains. 2) Similarly, to enhance the generalization ability of $\tilde{\mathbf{z}}^{I}_{i}$, we also utilize the $\mathcal{L}^{V(I)}_{kg}$ loss \cite{yao2023visuallanguage} to reduce the distance between $\tilde{\mathbf{z}}^{I}_{i}$ and $\mathbf{z}^{clip}_i$: $\mathcal{L}^{V(I)}_{kg}=||\tilde{\mathbf{z}}^{I}_{i}-\mathbf{z}^{clip}_i||_2^2$.

In addition, we also finetune the domain-specific visual encoder $\mathbf{f}_s(\cdot)$ to generate the prototype in Sec.~\ref{prototype} for subsequent steps. 
We employ the contrastive learning objective $\mathcal{L}^{V(S)}_{ce}$ calculated between $\tilde{\mathbf{z}}_i^S$ and $\tilde{\mathbf{h}}_m$ to guide the training of domain-specific visual prompts. 
Finally, the total loss function can be formulated as the above four terms:
\begin{equation} \label{loss_2}
\mathcal{L}^{V}_{all}=\mathcal{L}^{V(I)}_{ce}+\mathcal{L}^{V(S)}_{ce}+\alpha_2*\mathcal{L}^{V(I)}_{kg}+\beta_1*\mathcal{L}_{mix},
\end{equation}
where $\alpha_2, \beta_1$ are hyper-parameters to balance the loss terms.
In summary, the domain-invariant and domain-specific visual prompts are optimized using cross-entropy losses $\mathcal{L}_{ce}^{V(I)}$ and $\mathcal{L}_{ce}^{V(S)}$, regularized by normalization term $\mathcal{L}_{kg}^{V(I)}$ and confusion term $\mathcal{L}_{mix}$.
During the backward-propagation stage, the two visual prompts $\mathbf{E}_I$ and $\mathbf{E}_s$ are jointly updated by Eq.~\ref{loss_2}.

\subsection{Worst Explicit Representation Alignment}
\label{worst-case}
In this section, we introduce a simple yet effective Worst Explicit Representation Alignment module (WERA) to further enhance the invariance of the visual representations.
We decompose the simulation process into \emph{\textbf{two steps}}: 
\emph{\textbf{(1) Modeling the worst-case distribution}} for each training sample through a set of stylization prompts.
\emph{\textbf{(2) Aligning the explicit visual representation on worst-case distribution and source domain distribution}} in the feature space to fine-tune the domain-invariant visual prompt $\mathbf{E}_I$ pre-trained above.
The entire training process of WERA is described in Appendix (Aig.~1).

Directly modeling the worst-case distribution may distort the semantic content of images, since large perturbations in the feature space can change class-relevant information.
Therefore, we design a stylization strategy that only adjusts the statistical properties (mean and variance) of visual features.
This design keeps the semantic meaning unchanged while still allowing effective style transformations for distributional robustness.
Specifically, we partition the visual encoder into two components $\mathbf{f}^{I}_l(\cdot)$ and $\mathbf{f}^{I}_r(\cdot)$ to obtain the intermediate features. 
$\mathbf{f}^{I}_l(\cdot)$ represents the first linear layer of the visual encoder $\mathbf{f}_I(\cdot)$, while $\mathbf{f}^{I}_r(\cdot)$ comprises the remaining part of the visual encoder.
Let $\hat{\mathbf{z}}_i^I=\mathbf{f}^I_l(\mathbf{x}_i),\hat{\mathbf{z}}_i^I\in \mathbb{R}^{L\times C}$ be the output of the first linear layer, with L and C denoting the length of visual tokens and channel number, respectively. 
Then the stylized intermediate feature $\hat{\mathbf{z}}^w_i$ can be formulated as: 
\begin{equation} \label{function:IN}
\left \{
\begin{array}{ll}     \hat{\mathbf{z}}_i^w=\gamma(\hat{\mathbf{z}}_i^I-\mu(\hat{\mathbf{z}}_i^I))/\sigma(\hat{\mathbf{z}}_i^I)+\beta,
     \vspace{2.0mm} \\
     \mu(\hat{\mathbf{z}}_i^I)=\frac{1}{L}\sum\limits_{l=1}^{L}\hat{\mathbf{z}}_{i,l}^I,
     \vspace{2.0mm} \\
     \sigma^2(\hat{\mathbf{z}}_i^I)=\frac{1}{L}\sum\limits_{l=1}^{L}(\hat{\mathbf{z}}_{i,l}^I-\mu(\hat{\mathbf{z}}_i^I)),
\end{array} \right.
\end{equation}
where $\gamma,\beta\in \mathbb{R}^{C}$ are affine transformation parameters and $\hat{\mathbf{z}}_{i,l}^I\in \mathbb{R}^{C}$ denote the $l_{th}$ visual token of $\hat{\mathbf{z}}_{i}^I$.
Following the optimization objective of Wasserstein Distributionally Robust Learning in Eq.~\ref{function:worst-case}, 
we map the statistics (mean and variance) to the worst-case distribution.
Since the stylization strategy generates a new sample in the feature space by mixing the mean and variance, we first calculate the mixed mean $\hat{\mu}(\mathcal{A}_i,\cdot)$ and mixed variance $\hat{\gamma}(\mathcal{A}_i,\cdot)$ for all training samples
$\{(\mathbf{x}_i,y_i)\}_{i=1}^{N^S}$ at the beginning of the first epoch:
\begin{equation} \label{function:statistics_calculate}
\left \{
\begin{array}{ll}
\hat{\mu}(\mathcal{A}_i,\hat{\mathbf{z}}_i^I)=\alpha_{i,0}\mu(\hat{\mathbf{z}}_i^I)+\sum\limits_{j=1}^{M}\alpha_{i,j}\mu(\mathbf{f}_l(\mathbf{x}_j)),
\vspace{2.0mm} \\
\hat{\sigma}(\mathcal{A}_i,\hat{\mathbf{z}}_i^I)=\alpha_{i,0}\sigma(\hat{\mathbf{z}}_i^I)+\sum\limits_{j=1}^{M}\alpha_{i,j}\sigma(\mathbf{f}_l(\mathbf{x}_j)),
\end{array}\right.
\end{equation}
where $M$ is the number of bases used to construct the mean and variance of the worst-case distribution, and these bases are randomly selected from the images in the same batch as $\mathbf{x}_i$. 
A set of learnable prompts $\mathcal{A}_i=\{\alpha_{i,j}\}_{j=1}^M, \mathcal{A}_i\in \mathbb{R}^{M}$ in Eq.~\ref{function:statistics_calculate} controls the direction of the stylization.
We then apply mixed mean $\hat{\mu}(\mathcal{A}_i,\hat{\mathbf{z}}_i^I)$ and mixed std $\hat{\sigma}(\mathcal{A}_i,\hat{\mathbf{z}}_i^I)$ (for simplicity, we write $\hat{\mu}$ and $\hat{\sigma}$ meaning $\hat{\mu}(\mathcal{A}_i,\hat{\mathbf{z}}_i^I)$ and $\hat{\sigma}(\mathcal{A}_i,\hat{\mathbf{z}}_i^I)$) deviation to the intermediate features $\hat{\mathbf{z}}_i^I$ to obtain the stylized
intermediate feature $\hat{\mathbf{z}}^w_i$ by replacing $\gamma,\beta$ in Eq.~\ref{function:IN}: 
\begin{equation} \label{function:worst-output}
    \hat{\mathbf{z}}^w_{i}(\mathcal{A}_i,\hat{\mathbf{z}}^I_i)=\hat{\mu}+\hat{\sigma}(\hat{\mathbf{z}}_i^I-\mu(\hat{\mathbf{z}}_i^I))/\sigma(\hat{\mathbf{z}}_i^I).
\end{equation}

Then the stylized visual feature can be formulated as $\tilde{\mathbf{z}}_i^w=\mathbf{f}^I_r(\hat{\mathbf{z}}^w_{i};\mathbf{E}_I)$.
To obtain the worst-case training sample in the feature space, we first update stylized prompt $\mathcal{A}_i$ for $i_{th}$ intermediate feature $\tilde{\mathbf{z}}^w_i$ by maximizing the robust surrogate function $\mathcal{L}_{worst}$ with the same manner as Eq.~\ref{function:worst-case}, 
including the constraint loss term $\mathcal{L}^{W(I)}_{ce}$ and L2 distillation loss term $\mathcal{L}^{W(I)}_{kg}$.
The contrastive learning objective  $\mathcal{L}^{W(I)}_{ce}$ is calculated between $\tilde{\mathbf{z}}_{i}^w$ and fixed domain invariant embedding  $\tilde{\mathbf{w}}_k$: 
\begin{equation} \label{function:worst_probability}
    p^{w}_k(y_i \mid \tilde{\mathbf{z}}^{w}_{i}; \tilde{\mathbf{W}})=\frac{\exp (<\tilde{\mathbf{z}}^{w}_{i},\tilde{\mathbf{w}}_{k}> / \tau)}{\sum\limits_{k=1}^{N_{c}} \exp (<\tilde{\mathbf{z}}^{w}_{i}, \tilde{\mathbf{w}}_{k}> / \tau)},
\end{equation}
\begin{equation}
\label{function:worst_ce}
    \mathcal{L}^{W(I)}_{ce}=-\frac{1}{B} \sum_{i=1}^B \mathbf{y}_i \cdot \log \mathbf{P}^{w(I)}(y_i \mid \tilde{\mathbf{z}}^{w}_{i}; \tilde{\mathbf{W}}),
\end{equation}
where the $\mathbf{P}^{w(I)} \in \mathbb{R}^{N_c}$ in Eq.~\ref{function:worst_ce} is the collection of probabilities for classifying $\tilde{\mathbf{z}}^{w}_{i}$, and $p_k^w$ in Eq.~\ref{function:worst_probability} represents the probabilies of stylized visual feature belonging to $k_{th}$ category. 
The $\mathcal{L}^{W(I)}_{kg}$ is calculated between the stylized visual feature $\tilde{\mathbf{z}}^w_{i}$ and domain-invariant visual feature $\tilde{\mathbf{z}}^I_i$ in Eq.~\ref{image_feature}: $\mathcal{L}^{W(I)}_{kg}=\|\tilde{\mathbf{z}}^{w}_{i}-\tilde{\mathbf{z}}^{I}_{i}\|^{2}_{2}$.
Then the $\mathcal{L}_{worst}$ can be formulated as:
\begin{equation} \label{function:worst_all}
    \mathcal{L}_{worst}=\mathcal{L}^{W(I)}_{ce}-\gamma^{'}*\mathcal{L}^{W(I)}_{kg},
\end{equation}
where $\gamma^{'}$ in Eq.~\ref{function:worst_all} is the hyper-parameter that balances the degree of the penalty.
Maximizing Eq.~\ref{function:worst_all} can be interpreted as a finite-dimensional search within a Wasserstein ball centered on the source domain distribution, a process that may demand substantial computational resources.
Inspired by Adversarial attacks~\cite{ zhang2020principal}, to speed up the searching process, after the gradient back-propagation, we update the stylization prompt $\mathcal{A}_i$ using the gradient direction and fixed learning rate $\eta$:
\begin{equation} \label{function:worst-update}
\hat{\mathcal{A}}_i^{N_k}=\mathcal{A}^{N_{k-1}}_i+\eta* sign(\nabla_{\mathcal{A}}\mathcal{L}_{worst}(\tilde{\mathbf{z}}_{i}^w,\tilde{\mathbf{w}}_k)),  
\end{equation}
\begin{equation} \label{function:worst-normalized}
\alpha_{i,j}^{N_k}=\hat{\alpha}_{i,j}^{N_k}/\sum\limits_{j=1}^{M}\hat{\alpha}_{i,j}^{N_k},
    \ \forall \hat{\alpha}_{i,j}^{N_k} \in \hat{\mathcal{A}}_i^{N_k},
\end{equation}
where the normalized $\mathcal{A}_i^{N_k}$ is a set of stylized prompts $\{\alpha^{N_k}_{i,j}\}_{j=1}^{M}$ calculated in Eq.~\ref{function:worst-normalized} corresponds to $i_{th}$ intermediate feature $\hat{\mathbf{z}}_i^w$ in the $N_k$ iteration. 
We define $N_K$ as the total inner training iterations.
After $N_K$ iterations in inner optimization with Eq.~\ref{function:worst-update} and Eq.~\ref{function:worst-normalized}, we get $\mathcal{A}_i^{N_K}$. 
Then we can obtain the final worst-distribution-based visual features $\tilde{\mathbf{z}}_{i}^w$ using Eq.~\ref{function:statistics_calculate} and Eq.~\ref{function:worst-output} with fixed $\mathcal{A}_i^{N_K}$.

Subsequently, by aligning the visual features on worst-case distribution and source domain distribution in the feature space, we can fine-tune the domain-invariant visual prompt $\mathbf{E}_I$ by the following weighted formulation:
\begin{equation} \label{function:loss_rrl}
    \mathcal{L}^{W}_{all}=(1-\alpha_{3})*\mathcal{L}^{V(I)}_{ce}+\alpha_{3}*\mathcal{L}^{W(I)}_{ce}+\alpha_{2}*\mathcal{L}^{V(I)}_{kg},
\end{equation}
where $\mathcal{L}^{W(I)}_{ce}$ in Eq.~\ref{function:worst_ce} is calculated between the final worst-distribution-based visual features $\tilde{\mathbf{z}}_{i}^w$ and fixed domain invariant text embedding $\tilde{\mathbf{w}}_k$. 
Additionally, the rest two loss terms $\mathcal{L}^{V(I)}_{ce}$ and $\mathcal{L}^{V(I)}_{kg}$ are formulated as the same manner in Eq.~\ref{loss_2}.
$\alpha_2, \alpha_3$ are two hyper-parameters to balance the loss terms.
During the backward-propagation stage, we keep the domain-specific visual prompt $\mathbf{E}_s$ fixed and only update the domain-invariant visual prompt $\mathbf{E}_I$ by Eq.~\ref{function:loss_rrl}.

\subsection{Domain-Specific Prototype Learning} \label{prototype}
\begin{figure}[t]
\centering
\includegraphics[width=0.482\textwidth]{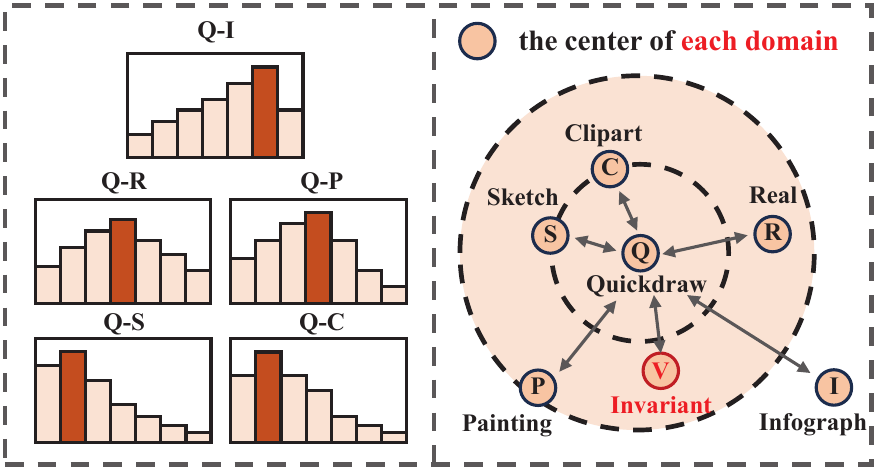}
\caption{
This figure demonstrates the distribution of distances between the quickdraw domain and the other five domains (sketch, clipart, infograph, painting, and real) within the same category in the DomainNet dataset. 
The left part shows the distance distributions of randomly selected positive pairs between quickdraw and the other five domains.
It is evident that quickdraw exhibits closer proximity to the sketch and clipart domains, while it is comparatively farther from the infograph domain.
It is important to highlight that the distributional distance between domain-invariant representation and quickdraw domain is larger than quickdraw domain and sketch domain.
}
\label{fig:example2}
\end{figure}

Inspired by the existing theory~\cite{dubey2021adaptive} on ensemble learning, 
a key limitation of the domain-invariant classifier is their potential divergence from the optimal target domain classifier, as compared to the classifier derived from individual source domains, as visualized in Fig.~\ref{fig:example2}.
To fully exploit the domain-specific information for DG, we further propose a new ensemble learning strategy: Domain-Specific Prototype Learning (DSPL) to devise a domain-specific predictor.
Given an unseen domain image, we can intuitively adopt relevance-inspired prototype prediction to well utilize the domain-specific visual features, which could reduce the domain shift between source and target data during inference.

\textbf{Prototype Initialization}. 
Given the pre-trained domain-specific visual encoder $\mathbf{f}_s(\cdot)$, we generate the prototype for each category within each specific domain. Specifically, given $N_c$ classes under $N_s$ source domains, we can obtain a prototype tensor: $\mathcal{C}\in \mathbb{R}^{N_c N_s\times d}$ as follows:
\begin{equation}
\mathcal{C}_{m,k,:} = \frac{1}{N^s_m}\sum_{\mathbf{x}_i\in \mathcal{D}^S_m,} \sum_{y_i=k} \mathbf{f}_s(\mathbf{x}_i; \mathbf{E}_s),
\label{PrototypeInit}
\end{equation}
where $d$ is the feature dimension, $y_i$ is the corresponding label of the input image $\mathbf{x}_i$, $N^s_m$ is the total number of instances in the $m_{th}$ source domain $\mathcal{D}_m^S$ with class label $y_i$. 
For simplicity, we re-write the prototype tensor $\mathcal{C}$ as a two-dimensional matrix as $\mathcal{C}\in \mathbb{R}^{N_c N_s\times d}$ in the following and store it in the memory bank during model training.

After constructing the prototype memory bank, we also transfer their corresponding labels into one-hot encoding as each prototype corresponds to one class label. 
Therefore, we can term the prototype learning as a key-value cache model~\cite{zhang2022tip}, where the key is the prototype for each class $\mathcal{C} \in \mathbb{R}^{N_c N_s\times d}$, and their corresponding value is the label set $\mathbf{L}_c\in \mathbb{R}^{N_c N_s \times N_c}$. During inference, given a test image $\mathbf{x}_i$ which serves as the query for retrieving from the cache model, we first extract the domain-specific visual features $\mathbf{f}_s(\mathbf{x}_i;\mathbf{E}_s) \in \mathbb{R}^{1\times d}$ by the visual encoder $\mathbf{f}_s(\cdot)$. Then the domain-specific prediction $\mathbf{P}^{S}_{i}$ can be calculated as follows:
\begin{equation}
    \mathbf{P}^{S}_{i}=\varphi (\mathbf{x}_i,\mathcal{C})\mathbf{L}_{c},
    \label{DS_Prediction}
\end{equation}
where $\varphi (\mathbf{x}_i, \mathcal{C})\in \mathbb{R}^{1\times N_c N_s}$ denotes the affinities between the query feature $\mathbf{f}_s(\mathbf{x}_i)$ and the prototypes $\mathcal{C}\in \mathbb{R}^{N_c N_s\times d}$ stored in the memory bank. It can be calculated as:
\begin{equation}
\varphi (\mathbf{x}_i,\mathcal{C})=\exp (-\beta(1-\mathbf{f}_s(\mathbf{x}_i;\mathbf{E}_s) \mathcal{C}^{\top})),
\end{equation}
where $\beta$ is a modulating hyper-parameter to control the sharpness of the similarity output. $\mathbf{f}_s(\mathbf{x}_i;\mathbf{E}_s) \mathcal{C}^{\top}$ can be viewed as the cosine similarity between the test visual feature $\mathbf{f}_s(\mathbf{x}_i;\mathbf{E}_s)$ and the prototypes $\mathcal{C}$ for all domain classes, as both of the key and query features are $L2$ normalized. After that, the domain-specific prediction based on the cache model can be obtained by the linear combination of the domain-specific cache values $\mathbf{L}_{c}$ weighted by query-key similarities, as illustrated in Eq.~\ref{DS_Prediction}.

Finally, to exploit the domain-invariant and specific information for DG, we make the final prediction by combining the domain-invariant and specific predictions as follows:
\begin{equation} \label{prototype_weight}
\mathbf{P}_{i} = \mathbf{P}^{I}_{i} + \beta_2*\mathbf{P}^{S}_{i},
\end{equation}
where $\mathbf{P}^{I}_{i}$ is the domain-invariant prediction and can be calculated as $\mathbf{P}^{I}_{i}=\mathbf{P}(y_i \mid \tilde{\mathbf{z}}_{i}^{I}; \tilde{\mathbf{W}})$ by utilizing the domain invariant visual features and the corresponding textual embeddings. $\beta_2$ is the trade-off parameter.

Besides, we further treat the prototypes $\mathcal{C}\in \mathbb{R}^{N_c N_s\times d}$ in the memory bank as learnable parameters with the mean features as initialization illustrated in Eq.~\ref{PrototypeInit}, then we finetune $\mathcal{C}$ via SGD for several epochs. Updating the prototypes in the memory bank can boost the estimation of affinities, which is able to calculate the cosine similarities between the test feature and the prototypes more accurately. In contrast, the values $\mathbf{L}_c$ are one-hot encodings representing the ground-truth annotations and should be kept frozen to well memorize the category information during training. 

\section{Theoretical Analysis} \label{sec:theoretical analysis}
Here we first provide the robustness guarantee for our proposed WERA in Sec.~\ref{sec:robustness_guarantee}, and then we derive the generalization bounds based on the covering number~\cite{vershynin2018high} for our proposed WERA in Sec.~\ref{sec:generalization_bounds}.
\subsection{Robustness Guarantee} \label{sec:robustness_guarantee}
Recall the objective of WERA is to optimize the empirical worst-case distribution $\hat{\mathcal{D}}^Q$ in a Wasserstein ball centered by empirical source domain distribution $\hat{\mathcal{D}}^S$. 
In this subsection, we derive the robustness guarantee for the WERA:
\begin{theorem}(Robustness Guarantee for WERA) \label{lemma_1}
Let $\hat{\mathcal{D}}^{S}$ be the empirical source domain distribution. 
We can define the empirical distribution set $\hat
{\mathcal{P}}^S=\{\hat{\mathcal{D}}:W_1(\hat{\mathcal{D}},\hat{\mathcal{D}}^S)\leq \rho\}$ with worst-case distribution $\hat{\mathcal{D}}^Q$ in a $\rho-$radius Wasserstein ball as:
\begin{equation} \label{function:distribution-set}
\hat{\mathcal{D}}^{Q}=\arg\max\limits_{\hat{\mathcal{P}}^S}\{\mathbb{E}_{\hat{D}} 
[\ell(\theta;\hat{\mathcal{D}})]-\gamma^{'} W_1(\hat{\mathcal{D}},\hat{\mathcal{D}}^S)\}.
\end{equation}
Then we denote the Wasserstein distance between worst-case distribution $\hat{\mathcal{D}}^{Q}$ and the source domain distribution $\hat{\mathcal{D}}^{S}$ as $\hat{\rho}^S=W_1(\hat{\mathcal{D}}^{Q},\hat{\mathcal{D}}^{S})$.
Let loss $\ell$ be convex, symmetric, bounded, and obey triangle equality.
Based on the existing theory~\cite{sinha2017certifying},
we can define a Wasserstein ball centered by $\hat{\mathcal{D}}^S$ as $\{\hat{\mathcal{D}}^{'}:W_1(\hat{\mathcal{D}}^{'},\hat{\mathcal{D}}^S)\leq \hat{\rho}^S\}\subset \hat{\mathcal{P}}^S$.
Then, for any $\gamma^{'}\geq 0$, we have:
\begin{equation} \label{function:robustness}
\begin{aligned}
\sup\limits_{\hat{\mathcal{D}}^{'}
}\mathbb{E}_{\hat{\mathcal{D}}^{'}}[\ell(\theta;\hat{\mathcal{D}}^{'})]
&=\mathbb{E}_{\hat{S}}
[h_{\gamma^{'}}(\theta;\hat{\mathcal{D}}^S)]+\gamma^{'}\hat{\rho}^S \\
&=\mathbb{E}_{\hat{Q}}[\ell(\theta;\hat{\mathcal{D}}^Q)]+\gamma^{'}\hat{\rho}^S, \\
where \ h_{\gamma^{'}}(\theta;\hat{\mathcal{D}}^S)&=\ell(\theta;\hat{\mathcal{D}}^S)-\gamma^{'}c(\hat{\mathcal{D}}^Q,\hat{\mathcal{D}}^S),
\end{aligned}  
\end{equation}
where the $c(\cdot,\cdot)$ in Eq.~\ref{function:robustness} is the continuous cost function defined in Wasserstein distance in Eq.~\ref{function:wasserstein-distance}. 
We can prove the theorem.~\ref{lemma_1} by choosing $\hat{\rho}^S$ as $\rho$ in theory~\cite{sinha2017certifying} Proposition \textcolor{red}{1}.
\end{theorem}
Theorem.~\ref{lemma_1} justifies that the optimization objective of WERA can exactly guarantee the distributional robustness inside a $\hat{\rho}^S-$radius ball, that is, given a proper $\gamma^{'}$, our proposed method will find a distribution $\hat{\mathcal{D}}^{Q}$, whose distance from the source domain distribution $\hat{\mathcal{D}}^{S}$ is $\hat{\rho}^S$, and we can guarantee that the learned $\hat{\mathcal{D}}^{Q}$ is exactly the worst-case distribution in the $\hat{\rho}^S-$radius ball centered at $\hat{\mathcal{D}}^{S}$. 

\subsection{Generalization Bounds} \label{sec:generalization_bounds}
In this subsection, we analyze the generalization bounds of the WERA optimization objective with the help of the usual covering number as notion of complexity.
\begin{definition} (Definition of covering number~\cite{vershynin2018high}).
Let $(\mathcal{F},\|\cdot\|)$ denote a metric space with respect to $\|\cdot\|$ defined on $\mathcal{F}$. 
We say that $\hat{\mathcal{F}}$ is a $\epsilon$ covering of $\mathcal{F}$, if for any element $\ell\in \hat{\mathcal{F}}$ there exists an element $\hat{\ell}\in \hat{\mathcal{F}}$ such that $\|\ell-\hat{\ell}\|\leq \epsilon$. Then the number of $\epsilon$ covering of $\hat{\mathcal{F}}$ is expressed as:
\begin{equation}
    \mathcal{N}(\mathcal{F},\epsilon,\|\cdot\|)=\min\{|\hat{\mathcal{F}}|:\hat{\mathcal{F}} \ is \ a \ \epsilon-covering \ of \ \mathcal{F}\}.
\end{equation}
\end{definition}
\begin{theorem} (Generalization Bounds for WERA)
\label{lemma_2}
Let $\hat{\mathcal{D}}^S$ be the empirical source domain distribution sampled from $\mathcal{D}^S$.  
For the model class $\mathcal{F}=\{\ell(\theta,\cdot):\theta\in \Theta\}$ equipped with the $L^{\infty}$ norm.
Assume $\ell(\theta;\cdot)\leq \mathcal{M}_{\ell}$ for all $\theta\in \Theta$.
We define the distribution set $\mathcal{P}^S=\{\mathcal{D}:W_1(\mathcal{D},\mathcal{D}^S)\leq \rho\}$.
Based on the theory~\cite{bartlett2002rademacher} Theorem \textcolor{red}{3} and definition of covering number, 
for a fixed $t>0$, with the probability at least $1-e^{-t}$, simultaneously for all $\gamma^{'}\geq 0$, $\rho\geq 0$, and some numerical constants $b_1,b_2>0$, we have:
\begin{equation}
\begin{aligned}
\sup\limits_{\mathcal{D}}\mathbb{E}_{\mathcal{D}}[\ell(\theta;\mathcal{D})]&\leq \mathbb{E}_{\hat{S}}[h_{\gamma^{'}}(\theta;\hat{\mathcal{D}}^S)]+\gamma^{'}\rho+\epsilon_{n}(t), \\
where \ \epsilon_{n}(t)&=b_{1}\mathcal{M}_{\ell}\sqrt{\frac{t}{\sum_{i=1}^{N_s}N_m^s}} \\
+\gamma^{'}b_{2}\sqrt{\frac{\mathcal{M}_{\ell}}{\sum_{i=1}^{N_s}N_m^s}}&\int_0^1\sqrt{\log \mathcal{N}(\mathcal{F},\mathcal{M}_{\ell}\epsilon,\|\cdot\|_{L^{\infty}})}d\epsilon. 
\end{aligned}
\end{equation}
Based on the Eq.~\ref{function:robustness}, if we set $\rho=\hat{\rho}^S$, with the probability at least $1-e^{-t}$, for all $\theta\in \Theta$:
\begin{equation}
\begin{aligned}
\sup\limits_{\mathcal{D}}\mathbb{E}_{\mathcal{D}}[\ell(\theta;\mathcal{D})]&\leq \mathbb{E}_{\hat{S}}[h_{\gamma^{'}}(\theta;\hat{\mathcal{D}}^S)]+\gamma^{'}\hat{\rho}^S+\epsilon_{n}(t) \\
&=\sup\limits_{\hat{\mathcal{D}}^{'}
}\mathbb{E}_{\hat{\mathcal{D}}^{'}}[\ell(\theta;\hat{\mathcal{D}}^{'})]+\epsilon_{n}(t).
\end{aligned}    
\end{equation}
\end{theorem}
Theorem.~\ref{lemma_2} justifies that the empirical optimization objective given by Eq.~\ref{function:distribution-set} can achieve consistent learning performance with the true optimization objective.
Under proper conditions, the bound can attain $\mathcal{O}(\sqrt{1/\sum_{i=1}^{N_s}N_m^s})$.

\begin{table*}[!htbp]
\small
\begin{center}
\caption{
Multi-source DG results (\%) on PACS, VLCS, OfficeHome, TerraInc, and DomainNet benchmark datasets. 
The best performances in comparisons are highlighted in \textbf{bold} and the second-best ones are marked with \underline{underlines}.
}
\label{tab:multi-DomainBed}
\renewcommand\arraystretch{1.05}
\scalebox{0.94}{
\begin{tabular}{l@{\hspace{16pt}}!{\vrule width0.5pt}@{\hspace{17pt}}c@{\hspace{17pt}}!{\vrule width0.5pt}@{\hspace{18pt}}c@{\hspace{19pt}}c@{\hspace{17pt}}c@{\hspace{16pt}}c@{\hspace{16pt}}c@{\hspace{16pt}}!{\vrule width0.5pt}>{\hspace{13pt}}>{\columncolor{gray!30}}c<{\hspace{13pt}}}
\hline
\textbf{Method} &\textbf{Venue} &\textbf{PACS} &\textbf{VLCS} &\textbf{OfficeHome} &\textbf{TerraInc}    &\textbf{DomainNet}  &\textbf{Avg.}  \\ 
\hline
\multicolumn{8}{c}{\textit{ResNet-50 Based Method.}} \\
\hline
DANN~\cite{ganin2016domain}       
& IJCAI'16    
& 83.60\tsb{\(\pm\)0.40} & 78.60\tsb{\(\pm\)0.40} & 65.90\tsb{\(\pm\)0.60} 
& 46.70\tsb{\(\pm\)0.50} & 38.30\tsb{\(\pm\)0.40} & 62.60 \\
MLDG~\cite{li2018learning}        
& AAAI'18     
& 84.90\tsb{\(\pm\)1.00} & 77.20\tsb{\(\pm\)0.40} & 66.80\tsb{\(\pm\)0.60} 
& 47.70\tsb{\(\pm\)0.90} & 41.20\tsb{\(\pm\)0.10} & 63.60 \\
GroupDRO~\cite{sagawa2019distributionally}    
& ICLR'20     
& 84.40\tsb{\(\pm\)0.80} & 76.70\tsb{\(\pm\)0.60} & 66.00\tsb{\(\pm\)0.70} 
& 43.20\tsb{\(\pm\)1.10} & 33.30\tsb{\(\pm\)0.20} & 60.70 \\
RSC~\cite{huang2020self}         
& ECCV'20     
& 85.20\tsb{\(\pm\)0.90} & 77.10\tsb{\(\pm\)0.50} & 65.50\tsb{\(\pm\)0.90} 
& 46.60\tsb{\(\pm\)1.00} & 38.90\tsb{\(\pm\)0.50} & 62.70 \\
VREx~\cite{krueger2021out}        
& ICML'21     
& 84.90\tsb{\(\pm\)0.60} & 78.30\tsb{\(\pm\)0.20} & 66.40\tsb{\(\pm\)0.60}
& 46.40\tsb{\(\pm\)0.60} & 33.60\tsb{\(\pm\)2.20} & 61.90 \\
Mixstyle~\cite{zhou2021domain}
& ICLR'21     
& 85.20\tsb{\(\pm\)0.30} & 77.90\tsb{\(\pm\)0.50} & 60.40\tsb{\(\pm\)0.20} 
& 44.00\tsb{\(\pm\)0.70} & 34.00\tsb{\(\pm\)0.10} & 60.30 \\
ERM~\cite{vapnik2013nature}
& ICLR'21     
& 85.50\tsb{\(\pm\)0.20} & 77.30\tsb{\(\pm\)0.40} & 66.50\tsb{\(\pm\)0.30} 
& 46.10\tsb{\(\pm\)1.80} & 43.80\tsb{\(\pm\)0.10} & 63.90 \\
SAM~\cite{li2018domain}
& ICLR'21     
& 85.80\tsb{\(\pm\)0.20} & 79.40\tsb{\(\pm\)0.10} & 69.60\tsb{\(\pm\)0.10} 
& 43.30\tsb{\(\pm\)0.70} & 44.30\tsb{\(\pm\)0.00} & 64.50 \\
SagNet~\cite{nam2021reducing}
& CVPR'21     
& 86.30\tsb{\(\pm\)0.20} & 77.80\tsb{\(\pm\)0.50} & 68.10\tsb{\(\pm\)0.10} 
& 48.60\tsb{\(\pm\)1.00} & 40.30\tsb{\(\pm\)0.10} & 64.20 \\
MIRO~\cite{cha2022domain}
& ECCV'22     
& 85.40\tsb{\(\pm\)0.40} & 79.00\tsb{\(\pm\)0.00} & 70.50\tsb{\(\pm\)0.40} 
& 50.40\tsb{\(\pm\)1.10} & 44.30\tsb{\(\pm\)0.20} & 65.90 \\
GSAM~\cite{zhuang2022surrogate}
& ICLR'22     
& 85.90\tsb{\(\pm\)0.10} & 79.10\tsb{\(\pm\)0.20} & 69.30\tsb{\(\pm\)0.00} 
& 47.00\tsb{\(\pm\)0.80} & 44.60\tsb{\(\pm\)0.20} & 65.10 \\
SAGM~\cite{wang2023sharpness}
& CVPR'23     
& 86.60\tsb{\(\pm\)0.20} & 80.00\tsb{\(\pm\)0.30} & 70.10\tsb{\(\pm\)0.20} & 48.80\tsb{\(\pm\)0.90} & 45.00\tsb{\(\pm\)0.20} & 66.10 \\
DomainDrop~\cite{guo2023domaindrop}
& ICCV'23     
& 87.90\tsb{\(\pm\)0.30} & 79.80\tsb{\(\pm\)0.30} & 68.70\tsb{\(\pm\)0.10} & 51.50\tsb{\(\pm\)0.40} & 44.40\tsb{\(\pm\)0.50} & 66.50 \\
GMDG~\cite{tan2024rethinking}
& CVPR'24     
& 85.60\tsb{\(\pm\)0.30} & 79.20\tsb{\(\pm\)0.30} & 70.70\tsb{\(\pm\)0.20} & 51.10\tsb{\(\pm\)0.90} & 44.60\tsb{\(\pm\)0.10} & 66.30 \\
SMOS~\cite{luo2024grounding}
& CVPR'24     
& 89.40\tsb{\(\pm\)0.30} & 79.80\tsb{\(\pm\)0.10} & 71.60\tsb{\(\pm\)0.10} & 55.40\tsb{\(\pm\)0.40} & 45.30\tsb{\(\pm\)0.00} & 68.30 \\
RES~\cite{huang2025representation}
& ECCV'24     
& 90.00\tsb{\(\pm\)0.30} & 79.80\tsb{\(\pm\)0.20} & 71.80\tsb{\(\pm\)0.30} & 51.40\tsb{\(\pm\)0.60} & 46.70\tsb{\(\pm\)0.20} & 67.90 \\
\hline
\multicolumn{8}{c}{\textit{VIT-B/16 Based Method.}} \\
\hline              
SWAD~\cite{cha2022domain}
& NIPS'21
& 91.30\tsb{\(\pm\)0.10} & 79.40\tsb{\(\pm\)0.40} & 76.90\tsb{\(\pm\)0.10} & 45.40\tsb{\(\pm\)0.50} & 51.70\tsb{\(\pm\)0.80} & 68.94 \\
CLIP~\cite{radford2021learning}
& -
& 96.20\tsb{\(\pm\)0.10} & 81.70\tsb{\(\pm\)0.10} & 82.00\tsb{\(\pm\)0.10} & 33.40\tsb{\(\pm\)0.10} & 57.50\tsb{\(\pm\)0.10} & 70.16 \\
ERM~\cite{vapnik1999nature} 
& ICLR'21
& 93.70\tsb{\(\pm\)0.10} & 82.70\tsb{\(\pm\)0.10} & 78.50\tsb{\(\pm\)0.10} & 52.30\tsb{\(\pm\)0.10} & 53.80\tsb{\(\pm\)0.10} & 72.20 \\
CoOp~\cite{Zhou_2022}  
& IJCV'22
& 96.20\tsb{\(\pm\)0.10} & 77.60\tsb{\(\pm\)0.20} & 83.90\tsb{\(\pm\)0.10} & 48.80\tsb{\(\pm\)0.10} & 59.80\tsb{\(\pm\)0.10} & 73.26 \\
MIRO~\cite{cha2022domain} 
& ECCV'22
& 95.60\tsb{\(\pm\)0.80} & 82.20\tsb{\(\pm\)0.30} & 82.50\tsb{\(\pm\)0.10} & 54.30\tsb{\(\pm\)0.40} & 54.00\tsb{\(\pm\)0.30} & 73.72 \\
SMA~\cite{arpit2022ensemble}
& NIPS'22 
& 92.10\tsb{\(\pm\)0.20} & 79.70\tsb{\(\pm\)0.20} & 78.10\tsb{\(\pm\)0.10} & 48.30\tsb{\(\pm\)0.70} & 55.90\tsb{\(\pm\)0.20} & 70.82 \\
SEDGE~\cite{li2022domain} 
& -
& 96.10\tsb{\(\pm\)0.10} & 82.20\tsb{\(\pm\)0.10} & 80.70\tsb{\(\pm\)0.20} & 56.80\tsb{\(\pm\)0.30} & 54.70\tsb{\(\pm\)0.10} & 74.10 \\
DUPRG~\cite{niu2022domainunified} 
& ICLR'23
& 97.10\tsb{\(\pm\)0.20} & 83.90\tsb{\(\pm\)0.50} & 83.60\tsb{\(\pm\)0.30} & 42.00\tsb{\(\pm\)1.30} & 59.60\tsb{\(\pm\)0.30} & 73.24 \\ 
DPL~\cite{zhang2022domain} 
& -
& 97.30\tsb{\(\pm\)0.20} & \underline{84.30\tsb{\(\pm\)0.40}} & 84.20\tsb{\(\pm\)0.20} & 52.60\tsb{\(\pm\)0.60} & 56.70\tsb{\(\pm\)0.10} & 75.02 \\
GESTUR~\cite{lew2023gradient} 
& ICCV'23
& 96.00\tsb{\(\pm\)0.00} & 82.80\tsb{\(\pm\)0.10} & 84.20\tsb{\(\pm\)0.10} & 55.70\tsb{\(\pm\)0.20} & 58.90\tsb{\(\pm\)0.10} & 75.52 \\
MaPLe~\cite{khattak2023maple}
& CVPR'23
& 96.50\tsb{\(\pm\)0.20} & 82.20\tsb{\(\pm\)0.20}
& 83.40\tsb{\(\pm\)0.00} & 50.20\tsb{\(\pm\)0.90}
& 59.50\tsb{\(\pm\)0.30} & 74.40 \\
SIMPLE$^{+}$~\cite{li2023simple}
& ICLR'23
& \textbf{99.00}\tsb{\(\pm\)0.10} 
& 82.70\tsb{\(\pm\)0.40}
& \textbf{87.70}\tsb{\(\pm\)0.40} 
& \underline{59.00\tsb{\(\pm\)0.60}}
& \underline{61.90\tsb{\(\pm\)0.50}} 
& \underline{78.10} \\
VL2V-SD~\cite{addepalli2024leveraging}
& CVPR'24
& 95.67\tsb{\(\pm\)0.56} & 82.67\tsb{\(\pm\)0.36}
& 85.44\tsb{\(\pm\)0.27} & 41.18\tsb{\(\pm\)0.74}
& 58.71\tsb{\(\pm\)0.11} & 72.73 \\
CLIP-LoRA~\cite{zanella2024low}
& CVPR'24
& 97.10\tsb{\(\pm\)0.00} 
& 83.10\tsb{\(\pm\)0.00}
& 83.90\tsb{\(\pm\)0.00} 
& 55.70\tsb{\(\pm\)0.00}
& 58.40\tsb{\(\pm\)0.00} 
& 75.60 \\
SPG~\cite{bai2025soft}
& ECCV'24
& 97.00\tsb{\(\pm\)0.50} & 82.40\tsb{\(\pm\)0.40}
& 83.60\tsb{\(\pm\)0.40} & 50.20\tsb{\(\pm\)1.20}
& 60.10\tsb{\(\pm\)0.50} & 74.70 \\
\hline
Ours (DPR)
& CVPR'24
& 97.45\tsb{\(\pm\)0.10} & 86.43\tsb{\(\pm\)0.30} & 86.13\tsb{\(\pm\)0.20} & 57.10\tsb{\(\pm\)0.20} & 62.05\tsb{\(\pm\)0.10} & 77.83 
\\
Ours (PADG)
& -
& \underline{97.80\tsb{\(\pm\)0.13}} 
& \textbf{86.72}\tsb{\(\pm\)0.21} 
& \underline{87.06\tsb{\(\pm\)0.18}} 
& \textbf{59.32}\tsb{\(\pm\)0.38} 
& \textbf{62.74}\tsb{\(\pm\)0.14} 
& \textbf{78.73} \\
\hline
\end{tabular}}
\end{center}
\end{table*}

\section{Experiments}
\subsection{Datasets and Evaluation Protocols}
\textbf{Datasets.} 
Following the existing works~\cite{gulrajani2020search}, we evaluate our proposed method on the five DomainBed benchmark datasets, including PACS~\cite{li2017deeper}, VLCS~\cite{li2017deeper}, OfficeHome~\cite{venkateswara2017deep}, DomainNet~\cite{peng2019moment}, and TerraIncognita~\cite{beery2018recognition}. 
These datasets cover diverse types of domain shifts—such as stylistic variations in DomainNet and environmental changes in TerraInc—providing comprehensive evaluation scenarios for domain generalization. Representative examples of different domain shifts are shown in the Appendix (Fig.~\textcolor{red}{1}).
The PACS contains 4 domains: art painting, cartoon, photo, and sketch, with 9991 images of 7 categories.
The VLCS contains 4 domains: Caltech101, LabelMe, SUN09, and VOC2007, with 10729 images of 5 categories.
The OfficeHome contains 4 domains: art, clipart, product, and real, with 15558 images of 65 categories.
The DomainNet contains 6 domains: sketch, clipart, painting, real, infograph, and quickdraw, with 586,575 images of 345 categories.
The TerraIncognita contains photographs of wild animals taken by camera traps at 4 locations: L100, L38, L43, and L46, with 24788 images of 10 categories.

\textbf{Evaluation Metrics.} 
In multi-source domain generalization experiments, it is customary to designate one domain as the target domain while considering the remaining domains as the source domains. 
To ensure the reliable results, we calculate the average performance across multiple experiments.

\subsection{Implementation Details}
We choose ViT-B/16 pre-trained on CLIP~\cite{radford2021learning} as a backbone network. 
To construct prompts for the text and visual encoders, we refer to the open-source implementations of CoOp and VPT. 
For PADG, we set the text and visual prompt lengths to 16 and 12, respectively. 
The hyper-parameter $\alpha_1$ in Eq.~\ref{loss_1} for GPT-Assist Text Disentanglement (GAT) is set to 8.0. 
The hyper-parameter $\alpha_2$, and $\beta_1$ in Eq.~\ref{loss_2} for Imgae-Disentanglement Guided by Text (IMT) are set to 2.0, and 0.8. 
Additionally, the hyper-parameter $\alpha_3$ in Eq.~\ref{function:loss_rrl} for Worst Explicit Representation Alignment (WERA) is set to 0.4.
The inner training epoch $N_K$ is set to $10$. 
The fixed learning rate $\eta$ in Eq.~\ref{function:worst-update} is set to 0.05.
The hyper-parameter $\gamma^{'}$ in Eq.~\ref{function:worst_all} is set to 8.0.
The hyper-parameter $\alpha_4$ in Eq.~\ref{prototype_weight} for Domain-Specific prototype Learning (DSPL) is set to 5.0. 
In the first training stage, the model is trained for 80 epochs for the Cross-Modal Prompt Disentanglement (CMD) module (40 epochs for GAT and 40 epochs for IMT).
In the second training stage, the model is trained for 40 epochs for the WERA module.
The SGD Optimizer is adopted with a weight decay of 5e-4. The initial learning rate is 5e-4 for WERA and 1e-3 for rest training stage.
To ensure the reliability of our proposed methods, we independently repeat all experiments three times and report the average results (Avg-acc) obtained from these trials.

\subsection{Comparison with State-of-the-Art Methods}
To demonstrate the efficiency of our proposed method, we compare our PADG with state-of-the-art (SOTA) methods, including the baseline method~\cite{vapnik2013nature}, the data augmentation based method~\cite{zhou2021domain, huang2025representation, yan2020improve, yun2019cutmix}, the domain-invariant representation based method~\cite{guo2023domaindrop, tan2024rethinking, luo2024grounding}, the feature disentanglement based method~\cite{nam2021reducing}, the distributionally robust optimization based method~\cite{sagawa2019distributionally, krueger2021out}, the gradient operation based method~\cite{huang2020self, lew2023gradient, arpit2022ensemble}, the meta-learning based method~\cite{li2018learning}, the flatness-aware based method~\cite{li2018domain, zhuang2022surrogate, cha2022domain, wang2023sharpness, shinunknown}, and the Oracle model based method~\cite{cha2022domain, radford2021learning, Zhou_2022, niu2022domainunified, li2022domain, zhang2022domain, khattak2023maple, addepalli2024leveraging, zanella2024low, bai2025soft}.
The results are shown in Tab.~\ref{tab:multi-DomainBed}.

Notably, our proposed PADG achieves the highest average performance and sets new state-of-the-art (SOTA) results at 97.80\%, 86.72\%, 87.06\%, 59.01\%, and 62.74\% on all benchmark datasets.
Especially in the challenge case with large style discrepancies, such as DomainNet, our PADG can even gain significant performance than SPG by 2.64\%.
Meanwhile, it is important to highlight that the representative VFM based method, \emph{e.g.} CLIP and CoOp, show limited performance with ERM on TerraInc (52.30\%$\rightarrow$33.40 and 52.30\%$\rightarrow$48.80) dataset.
We speculate that this may result from some rare categories in TerraInc (e.g., bobcat) that share similar semantic information with more common categories (e.g., cat).
This leads to the text encoder outputting similar textual embeddings, resulting in overlapping classification boundaries.
Notably, our proposed PADG substantially improves the ERM of $6.71$ and exceeds the existing SOTA method CLIP-LORA by $3.31\%$ on TerraInc.
This observation indicates that compared to the existing VFM based method, our PADG not only alleviates the negative impacts of the style shifts but also guarantees a provable generalization performance under varying viewpoints and background conditions.

To sum up, our proposed method offers four major advantages:
(1) The proposed PADG framework substantially improves model generalization under heterogeneous domain shifts (\emph{e.g.}, stylization and viewpoint changes) by introducing a novel prompt-based disentanglement mechanism;
(2) By incorporating the Wasserstein Distributionally Robust Learning principle, PADG generates diverse stylized domains and aligns visual representations between the original and perturbed distributions in the feature space, effectively mitigating distributional discrepancies between source and target domains;
(3) PADG jointly exploits domain-invariant and domain-specific knowledge through a dynamic prototype-based ensemble strategy, leading to more reliable and robust inference;
(4) Compared with existing DG methods, PADG provides a lightweight yet effective fine-tuning framework that achieves superior generalization performance with minimal computational overhead.
Overall, PADG successfully leverages the generalization ability of large-scale pre-trained models and consistently validates the effectiveness of our hypothesis across various benchmark datasets.

\begin{table}[!t]
\begin{center}
\caption{
\footnotesize{
SDG results (\%) on PACS. 
One domain is used as the source domain and the others are used as the target domain.
The best comparison performances are highlighted in \textbf{bold} and the second-best ones are marked with \underline{underlines}.
}}
\renewcommand\arraystretch{1.1}
\label{tab:multi-source-pacs}
\scalebox{0.89}{
\begin{tabular}{l|c|cccc|>{\columncolor{gray!30}}c}
\hline
\textbf{Method}  &\textbf{Venue} &\textbf{A}    &\textbf{C}      &\textbf{P}     &\textbf{S}     &\textbf{Avg.}   \\ 
\hline
DANN~\cite{ganin2016domain}       
& IJCAI'16    
& 79.00 & 76.50 & 48.70 & 57.90 & 65.50 \\
MMD~\cite{li2018domain}
& ICCV'18     
& 75.40 & 80.10 & 45.20 & 58.20 & 64.70 \\
Mixup~\cite{yan2020improve}
& -           
& 77.40 & 80.00 & 47.30 & 58.20 & 65.70 \\
CutMix~\cite{yun2019cutmix}
& CVPR'19     
& 71.10 & 76.40 & 37.70 & 50.40 & 58.90 \\
GroupDRO~\cite{sagawa2019distributionally}    
& ICLR'20     
& 79.00 & 79.00 & 42.00 & 60.80 & 65.20 \\
MTL~\cite{blanchard2021domain}
& JMLR'21     
& 76.70 & 78.70 & 44.70 & 59.50 & 64.90 \\
ARM~\cite{zhang2021adaptive}
& NeurIPS'21  
& 76.20 & 75.50 & 45.20 & 61.90 & 64.70 \\
VREx~\cite{krueger2021out}        
& ICML'21     
& 75.30 & 80.20 & 44.90 & 56.80 & 64.30 \\
Mixstyle~\cite{zhou2021domain}
& ICLR'21     
& 78.10 & 78.80 & 56.10 & 54.70 & 66.90 \\
ERM~\cite{vapnik2013nature}
& ICLR'21     
& 79.90 & 79.90 & 48.10 & 59.60 & 66.90 \\
SAM~\cite{li2018domain}
& ICLR'21     
& 77.70 & 80.50 & 46.70 & 54.20 & 64.80 \\
SagNet~\cite{nam2021reducing}
& CVPR'21     
& 77.40 & 78.90 & 47.60 & 56.40 & 65.10 \\
Fishr~\cite{rame2022fishr}
& ICML'22     
& 75.90 & 81.10 & 46.90 & 57.20 & 65.30 \\
RIDG~\cite{chen2023domain}
& ICCV'23     
& 76.20 & 80.00 & 48.50 & 54.80 & 64.90 \\
SAGM~\cite{wang2023sharpness}
& CVPR'23     
& 78.90 & 79.80 & 44.70 & 55.60 & 64.80 \\
ITTA~\cite{chen2023improved}
& CVPR'23     
& 78.40 & 79.80 & 56.50 & 60.70 & 68.80 \\
UDIM~\cite{shinunknown}
& ICLR'24      
& 82.40 & 84.20
& 68.80 & 64.00 
& 74.90 \\ 
\hline
CLIP~\cite{radford2021learning}
& -
& 87.57 & \underline{86.96} 
& \underline{86.11} & \underline{94.13} 
& \underline{88.69} \\
CoOp~\cite{Zhou_2022}
& IJCV'22
& \underline{88.28} & 85.24 
& 72.13 & 73.99 
& 79.91 \\
\hline
\textbf{Ours (PADG)}
& -
& \textbf{90.60} & \textbf{87.75} 
& \textbf{88.09} & \textbf{95.00} 
& \textbf{90.36} \\
\hline\end{tabular}}
\end{center}
\vspace{-4.5mm}
\end{table}

\begin{table}[t]
\begin{center}
\caption{
SDG (\%) on OfficeHome. 
One domain is used as the source domain and the others are used as the target domain.
The best comparison performances are highlighted in \textbf{bold}, and the second-best ones are marked with \underline{underlines}.
}
\renewcommand\arraystretch{1.1}
\label{tab:multi-source-officehome}
\scalebox{0.89}{
\begin{tabular}{l|c|cccc|>{\columncolor{gray!30}}c}
\hline
\textbf{Method}  &\textbf{Venue}    &\textbf{A}    &\textbf{C}      &\textbf{P}        &\textbf{R}    &\textbf{Avg.}  \\ 
\hline
DANN~\cite{ganin2016domain}       
& IJCAI'16    & 55.20 & 49.30  & 48.40  & 58.40 & 52.80 \\
MMD~\cite{li2018domain}
& ICCV'18     & 55.10 & 52.00  & 50.30  & 59.30 & 54.20 \\
Mixup~\cite{yan2020improve}
& -           & 55.50 & 54.10  & 49.40  & 59.40 & 54.60 \\
CutMix~\cite{yun2019cutmix}
& CVPR'19     & 53.50 & 52.20  & 47.70  & 60.20 & 53.40 \\
GroupDRO~\cite{sagawa2019distributionally}    
& ICLR'20     & 55.10 & 52.00  & 50.30  & 59.30 & 54.20 \\
MTL~\cite{blanchard2021domain}
& JMLR'21     & 55.30 & 53.30  & 49.00  & 60.40 & 54.50 \\
ARM~\cite{zhang2021adaptive}
& NeurIPS'21  & 55.00 & 51.60  & 47.30  & 59.30 & 53.30 \\
VREx~\cite{krueger2021out}        
& ICML'21     & 55.50 & 52.60  & 49.10  & 59.30 & 54.10 \\
Mixstyle~\cite{zhou2021domain}
& ICLR'21     & 44.30 & 29.80  & 33.60  & 48.50 & 39.00 \\
ERM~\cite{vapnik2013nature}
& ICLR'21     & 55.60 & 52.80  & 50.30  & 59.40 & 54.50 \\
SAM~\cite{li2018domain}
& ICLR'21     & 56.90 & 53.80  & 50.90  & 61.50 & 55.80 \\
SagNet~\cite{nam2021reducing}
& CVPR'21     & 56.90 & 53.40  & 50.80  & 61.20 & 55.60 \\
Fishr~\cite{rame2022fishr}
& ICML'22     & 55.10 & 51.20  & 49.20  & 59.90 & 53.90 \\
RIDG~\cite{chen2023domain}
& ICCV'23     & 56.80 & 55.40  & 50.50  & 60.90 & 55.90 \\
SAGM~\cite{wang2023sharpness}
& CVPR'23     & 57.70 & 54.80  & 51.50  & 61.40 & 56.30 \\
ITTA~\cite{chen2023improved}
& CVPR'23     
& 56.00 & 51.50  
& 50.50 & 61.60 
& 54.90 \\
UDIM~\cite{shinunknown}
& ICLR'24      
& 58.50 & 55.70  
& 54.50 & 64.50 
& 58.30 \\ 
\hline
CLIP~\cite{radford2021learning}
& -
& \underline{71.89} & \underline{79.71} 
& \underline{68.03} & 67.81 
& \underline{71.86} \\
CoOp~\cite{Zhou_2022}
& IJCV'22
& 69.32 & 72.25 
& 62.62 & \underline{68.50} 
& 68.17 \\
\hline              
\textbf{Ours (PADG)}
& -
& \textbf{75.37} & \textbf{81.97} 
& \textbf{71.04} & \textbf{71.99} 
& \textbf{75.09} \\
\hline\end{tabular}}
\end{center}
\vspace{-4.5mm}
\end{table}

\subsection{Results on Single Domain Generalization}
Following the existing work~\cite{shinunknown}, we evaluate the effectiveness of our proposed PADG using only the GPT-Assist Text Disentanglement module for a challenge case in DG: single domain generalization (SDG).
SDG focuses on generalizing a model learned from \textbf{only one} source domain to multiple unseen target domains.
We apply the ResNet-50 pretrained on CLIP as the backbone, while maintaining all other experimental settings consistent with those utilized in the multi-domain generalization.
Following~\cite{shinunknown}, we choose the PACS and OfficeHome as the benchmark datasets. 
The results are shown in Tab.~\ref{tab:multi-source-pacs} and Tab.~\ref{tab:multi-source-officehome}.
When the training domains are limited or even restricted to a single domain, it becomes challenging to extract robust and generalizable domain-invariant features.
Notably, our proposed method significantly outperforms the state-of-the-art method UDIM by 15.46\% and 16.79\%, respectively.
Especially when taking "Sketch" (PACS), and "Clipart" (OfficeHome) as the source domain, our PADG leads to 31.00\% and 26.27\% improvements, respectively. 
Furthermore, our proposed method demonstrates superior performance compared to methods utilizing large-scale pre-trained models (e.g., CLIP and CoOp), achieving improvements of 1.67\% and 10.45\% on PACS and 3.23\% and 6.92\% on OfficeHome, respectively.
It can be observed that the prompt tuning method exhibits limited performance in the single domain generalization setting, primarily due to the substantial domain discrepancy.
However, our proposed method fully leverage the powerful LLM to enforce consistency between the domain-invariant descriptions and learnable prompt, resulting the significant performance gain.
Overall, the encouraging results demonstrate the superiority of our method in alleviating the negative impacts of distributional discrepancy between different domains despite only one source domain being available for training, further verifying the effectiveness of text disentanglement.

\begin{table*}[!htbp]
\caption{Ablation study on individual components of our method on all benchmark datasets.
}
\centering
\renewcommand\arraystretch{1.1}
\scalebox{1.125}{
\footnotesize
\begin{tabular}{c!{\vrule width0.5pt}cccc!{\vrule width0.5pt}c@{\hspace{15pt}}c@{\hspace{12pt}}c@{\hspace{12pt}}c@{\hspace{13pt}}c!{\vrule width0.5pt}>{\hspace{8pt}}>{\columncolor{gray!30}}c<{\hspace{8pt}}}
\hline
\multirow{2}{*}{\textbf{Idx}}
& \multicolumn{4}{c!{\vrule width0.5pt}}{\textbf{Components}} 
& \multirow{2}{*}{\textbf{PACS}} 
& \multirow{2}{*}{\textbf{VLCS}} 
& \multirow{2}{*}{\textbf{OfficeHome}} 
& \multirow{2}{*}{\textbf{TerraInc}}  
& \multirow{2}{*}{\textbf{DomainNet}}  
& \multirow{2}{*}{\textbf{Avg.}} \\
\cline{2-5}
& \textbf{GAT}  &\textbf{IMT}  &\textbf{WERA}       & \textbf{DSPL} &~ &~ &~ &~ &~ \\
\hline
\multicolumn{11}{l}{\textit{Text-Disentanglement only.}} \\
\hline
1 & \textendash & \textendash 
& \textendash & \textendash 
& 96.20\tsb{\(\pm\)0.10} & 80.80\tsb{\(\pm\)0.23} 
& 83.90\tsb{\(\pm\)0.10} & 46.80\tsb{\(\pm\)0.10}  
& 59.50\tsb{\(\pm\)0.10} & 73.44 \\
2 & \checkmark & \textendash 
& \textendash & \textendash
& 96.52\tsb{\(\pm\)0.11} & 82.69\tsb{\(\pm\)0.10}
& 85.61\tsb{\(\pm\)0.08} & 47.98\tsb{\(\pm\)0.53} 
& 61.50\tsb{\(\pm\)0.04} & 74.86 \\
\hline
\multicolumn{11}{l}{\textit{Image-Disentanglement only.}} \\
\hline
3 & \textendash & \checkmark  
& \textendash & \textendash
& 96.90\tsb{\(\pm\)0.28} & 85.27\tsb{\(\pm\)0.35} 
& 85.54\tsb{\(\pm\)0.53} & 55.39\tsb{\(\pm\)0.61} 
& 60.34\tsb{\(\pm\)0.19} & 76.69 \\
4 & \textendash & \checkmark  
& \checkmark  & \textendash
& 97.48\tsb{\(\pm\)0.22} & 85.73\tsb{\(\pm\)0.17} & 86.01\tsb{\(\pm\)0.27} & 57.04\tsb{\(\pm\)0.32} & 61.18\tsb{\(\pm\)0.15} & 77.49 \\
5 & \textendash & \checkmark  
& \textendash & \checkmark
& 97.24\tsb{\(\pm\)0.33} & 86.09\tsb{\(\pm\)0.29} & 85.73\tsb{\(\pm\)0.49} & 56.04\tsb{\(\pm\)0.70}
& 60.81\tsb{\(\pm\)0.22} & 77.18 \\
6 & \textendash & \checkmark  
& \checkmark  & \checkmark
& 97.60\tsb{\(\pm\)0.17} & 86.31\tsb{\(\pm\)0.21}
& 86.12\tsb{\(\pm\)0.14} & 57.56\tsb{\(\pm\)0.26}
& 61.45\tsb{\(\pm\)0.08} & 77.81 \\
\hline
\multicolumn{11}{l}{\textit{Image and Text-Disentanglement.}} \\
\hline
7 & \checkmark & \checkmark  
& \textendash & \textendash
& 97.12\tsb{\(\pm\)0.11} & 85.67\tsb{\(\pm\)0.10}  
& 86.09\tsb{\(\pm\)0.10} & 56.82\tsb{\(\pm\)0.10}  
& 61.84\tsb{\(\pm\)0.10} & 77.51 \\
8 & \checkmark & \checkmark  
& \checkmark  & \textendash
& 97.57\tsb{\(\pm\)0.11} & 86.32\tsb{\(\pm\)0.17}
& 86.77\tsb{\(\pm\)0.11} & 58.35\tsb{\(\pm\)0.20} 
& 62.53\tsb{\(\pm\)0.15} & 78.28 \\
9 & \checkmark & \checkmark  
& \textendash & \checkmark
& 97.48\tsb{\(\pm\)0.10} & 86.50\tsb{\(\pm\)0.30}  
& 86.31\tsb{\(\pm\)0.20} & 57.50\tsb{\(\pm\)0.20} & 62.05\tsb{\(\pm\)0.10} & 77.97 \\
10 & \checkmark & \checkmark  
& \checkmark  & \checkmark 
& 97.80\tsb{\(\pm\)0.13} & 86.72\tsb{\(\pm\)0.21}  
& 87.06\tsb{\(\pm\)0.18} & 59.32\tsb{\(\pm\)0.38}  & 62.74\tsb{\(\pm\)0.14} & 78.73 \\ 
\hline
\end{tabular}
}
\setlength{\belowcaptionskip}{2pt}
\label{tab:2_analysis_abolution}
\end{table*}
\begin{table}[t]
\caption{
\footnotesize{
Analysis of the influence of the parameter $L_a$ on text disentanglement on VLCS (VL), OfficeHome (OH), and DomainNet (DN) datasets. Here, $L_a$ represents the number of responses generated by GPT-3 for each input question.
All experiments based on the $1^{st}$ line of Tab.~\ref{tab:2_analysis_abolution}.
$\Delta$: the difference between the current and the preceding line. 
}}
\centering	
\renewcommand\arraystretch{1.05}
\scalebox{0.97}{
\footnotesize
\begin{tabular}{c!{\vrule width0.5pt}l!{\vrule width0.5pt}c@{\hspace{18pt}}c@{\hspace{16pt}}c!{\vrule width0.5pt}>{\columncolor{gray!30}}c}
\hline
\multicolumn{2}{c!{\vrule width0.5pt}}{\textbf{Module}} 
& \textbf{VL} 
& \textbf{OH} 
& \textbf{DN} 
& \textbf{Avg.} \\
\hline
\multicolumn{2}{l!{\vrule width0.5pt}}{Baseline} 
& 80.80 & 83.90 & 59.50 & 74.73 \\
\multicolumn{2}{l!{\vrule width0.5pt}}{+ hand-craft description} 
& 81.37 & 84.59 & 60.71 & 75.56 \\
\multicolumn{2}{l!{\vrule width0.5pt}}{$\Delta$}
& \textcolor{red}{+0.57}
& \textcolor{red}{+0.69}
& \textcolor{red}{+1.21}
& \textcolor{red}{+0.83} \\
\hline
\multirow{3}{*}{$L_a$ = 1} 
& + fine-grained
& 81.54 & 84.68 & 60.82 & 75.68 \\
 
& + domain-invariant
& 81.58 & 84.65 & 60.75 & 75.66 \\
& $\Delta$
& \textcolor{red}{+0.04}
& \textcolor{green}{-0.03}
& \textcolor{green}{-0.07}
& \textcolor{green}{-0.02} \\
\hline
\multirow{3}{*}{$L_a$ = 3} 
& + fine-grained 
& 81.92 & 84.94 & 60.94 & 75.93 \\

& + domain-invariant
& 82.04 & 85.01 & 61.15 & 76.07 \\
& $\Delta$
& \textcolor{red}{+0.12}
& \textcolor{red}{+0.07}
& \textcolor{red}{+0.21}
& \textcolor{red}{+0.14} \\
\hline
\multirow{3}{*}{$L_a$ = 5} 
& + fine-grained  
& 82.15 & 85.08 & 61.23 & 76.15 \\

& + domain-invariant
& 82.37 & 85.32 & 61.38 & 76.36 \\
& $\Delta$
& \textcolor{red}{+0.22}
& \textcolor{red}{+0.24}
& \textcolor{red}{+0.15}
& \textcolor{red}{+0.21} \\
\hline
\multirow{3}{*}{$L_a$ = 7} 
& + fine-grained  
& 82.31 & 85.17 & 61.33 & 76.27 \\
 
& + domain-invariant
& 82.55 & 85.46 & 61.51 & 76.51 \\
& $\Delta$
& \textcolor{red}{+0.24}
& \textcolor{red}{+0.29}
& \textcolor{red}{+0.18}
& \textcolor{red}{+0.24} \\
\hline
\multirow{3}{*}{$L_a$ = 9} 
& + fine-grained  
& 82.38 & 85.29 & 61.41 & 76.36 \\
 
& + domain-invariant
& 82.69 & 85.61 & 61.50 & 76.60 \\
& $\Delta$
& \textcolor{red}{+0.31}
& \textcolor{red}{+0.32}
& \textcolor{red}{+0.09}
& \textcolor{red}{+0.24} \\
\hline
\multirow{3}{*}{$L_a$ = 11} 
& + fine-grained  
& 82.41 & 85.28 & 61.39 & 76.36 \\
 
& + domain-invariant
& 82.66 & 85.63 & 61.53 & 76.61 \\
& $\Delta$
& \textcolor{red}{+0.25}
& \textcolor{red}{+0.35}
& \textcolor{red}{+0.14}
& \textcolor{red}{+0.25} \\
\hline
\end{tabular}
}
\label{tab3:compare with other Guided-prompt}
\end{table}
\subsection{Ablation Study}
To validate the effectiveness of each component of our method, we conduct an ablation experiment on all DomainBed benchmark datasets in Tab.~\ref{tab:2_analysis_abolution}, respectively. 
\textbf{GAT} indicates we apply GPT-Assist text disentanglement sub-module in Sec.~\ref{Text_side}.
\textbf{IMT} denote our Image-Disentanglement Guided by Text sub-module in Sec.~\ref{Image_side}.
\textbf{WERA} denote our Worst Explicit Representation Alignment module in Sec.~\ref{worst-case}.
\textbf{DSPL} denotes our Domain-Specific Prototype Learning module in Sec.~\ref{prototype}.
Our baseline (see $1^{st}$ row) for text disentanglement involves CoOp with learnable domain-invariant and domain-specific text prompts but does not incorporate the guidance of GPT-Assist disentangled descriptions.

In addition to mean performance, we report the $t$-test and corresponding $p$-values to evaluate the statistical significance of performance gains, where smaller $p$-values indicate more reliable improvements ($p$~$<$~0.05 for significance, $p$~$<$~0.01 for high significance). The detailed computation procedure for the $t$-test and corresponding $p$-values is provided in the Appendix.

\begin{figure*}[t]
    \centering
    \includegraphics[width=0.98\textwidth]{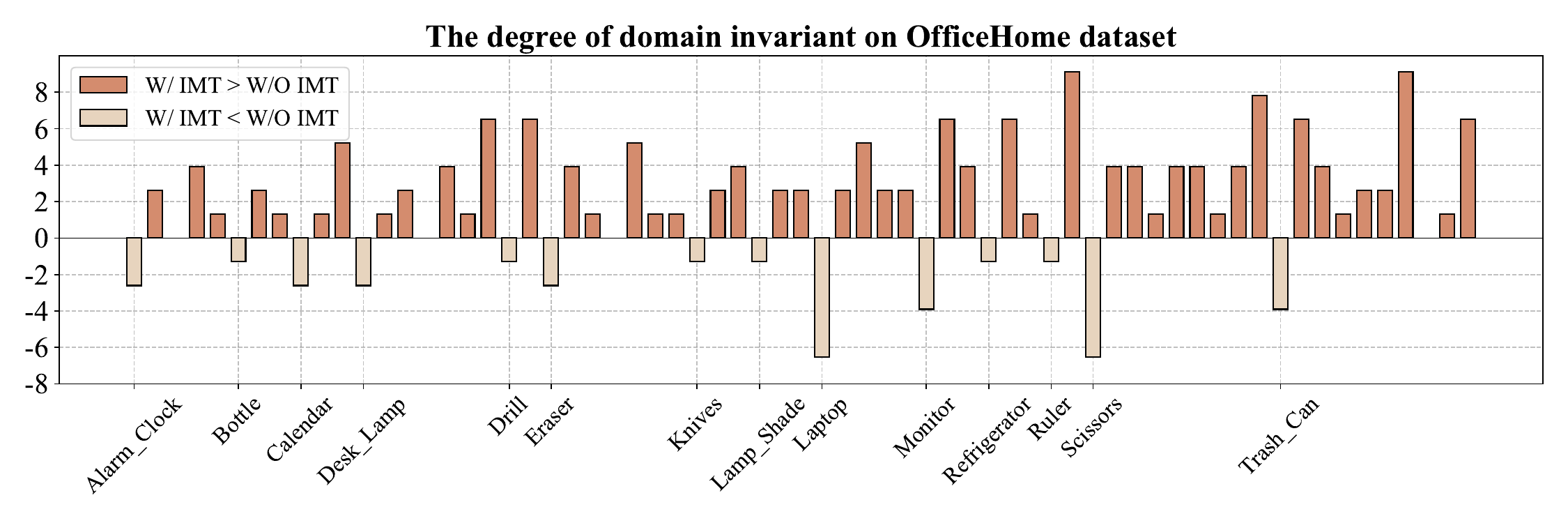}
    \vspace{-4.5mm}
    \caption{
    Comparisons of average source-target divergences ($\times 1000$) between the proposed IMT sub-module and the baseline without the IMT sub-module on the OfficeHome dataset.
    A lower divergence indicates a higher degree of domain invariance.
    We compute the difference in average divergence between the scenarios with and without the IMT sub-module (W/ IMT - W/O IMT). 
    The results indicate that our proposed IMT reduces divergence across most categories, thereby improving the degree of domain invariance and achieving the disentanglement of visual representation.
    }
    \label{pic:invariant_representation_officehome} 
\end{figure*}

\textbf{The effectiveness of GPT-Assist Text-Disentanglement (GAT).} 
Our GAT sub-module achieves improvements of 0.32\%, 1.89\%, 1.71\%, 1.18\%, and 2.00\% over the baseline (see $1^{st}$ and $2^{nd}$ rows in Tab.~\ref{tab:2_analysis_abolution}), with corresponding $t$-values of 3.72, 13.05, 23.12, 3.78, and 32.16 and two-tailed $p$-values of 0.020, $<$0.0001, $<$0.0001, 0.019, and $<$0.0001, all below 0.05, confirming significant improvements across all datasets.
When combined with the IMT sub-module, GAT further contributes to gains of 0.22\%, 0.40\%, 0.55\%, 1.43\%, and 1.50\% (see $3^{rd}$ and $7^{th}$ rows in Tab.~\ref{tab:2_analysis_abolution}), with $p$-values of 0.27, 0.21, 0.19, 0.018, and $<$0.001, showing that the improvements are more evident on complex datasets such as TerraInc and DomainNet.
This observation indicates that the text modality has rich semantic information and can easily be disentangled.
Meanwhile, as mentioned above, the original CLIP with template text input obtains a relatively slight performance gain on TerraInc.
However, our GAT module fully harnesses the potential of LLM to generate sufficient domain-invariant descriptions corresponding to each class, which enables the text classifier to more effectively differentiate between similar categories. 
With the guidance of disentangled domain-invariant and domain-specific text embeddings, we mitigate the distributional discrepancy between different domains in visual representation, resulting in a generalization gain.

We further investigate the effect of the number of fine-grained text descriptions $L_a$ generated by LLM on disentanglement, as shown in Tab.~\ref{tab3:compare with other Guided-prompt}.
Compared with the rigid template text input (see $2^{nd}$ row and $11^{th}$ row in Tab.~\ref{tab3:compare with other Guided-prompt}),  our fine-grained descriptions and domain-invariant descriptions bring significant performance improvement by 0.80\%, and 1.04\% on Avg-acc respectively, indicating that the simple template is insufficient to describe all categories.
Additionally, as $L_a$ increases, the performance of applying domain-invariant descriptions improves, with Avg-acc rising from 76.07\% to 76.61\%.
This consistently outperforms (see $\Delta$ in Tab.~\ref{tab3:compare with other Guided-prompt}) the use of fine-grained descriptions excepted $L_a=1$, which yields an increase from 75.93\% to 76.36\%.
These results further demonstrate that more diverse fine-grained descriptions are beneficial for the LLM in summarizing more distinctive domain-invariant descriptions.
Moreover, this observation also suggests that the domain-specific descriptions may hinder the ability of text prompt learning domain-invariant information.

\begin{figure}[t]
    \centering
    \includegraphics[width=0.49 \textwidth]{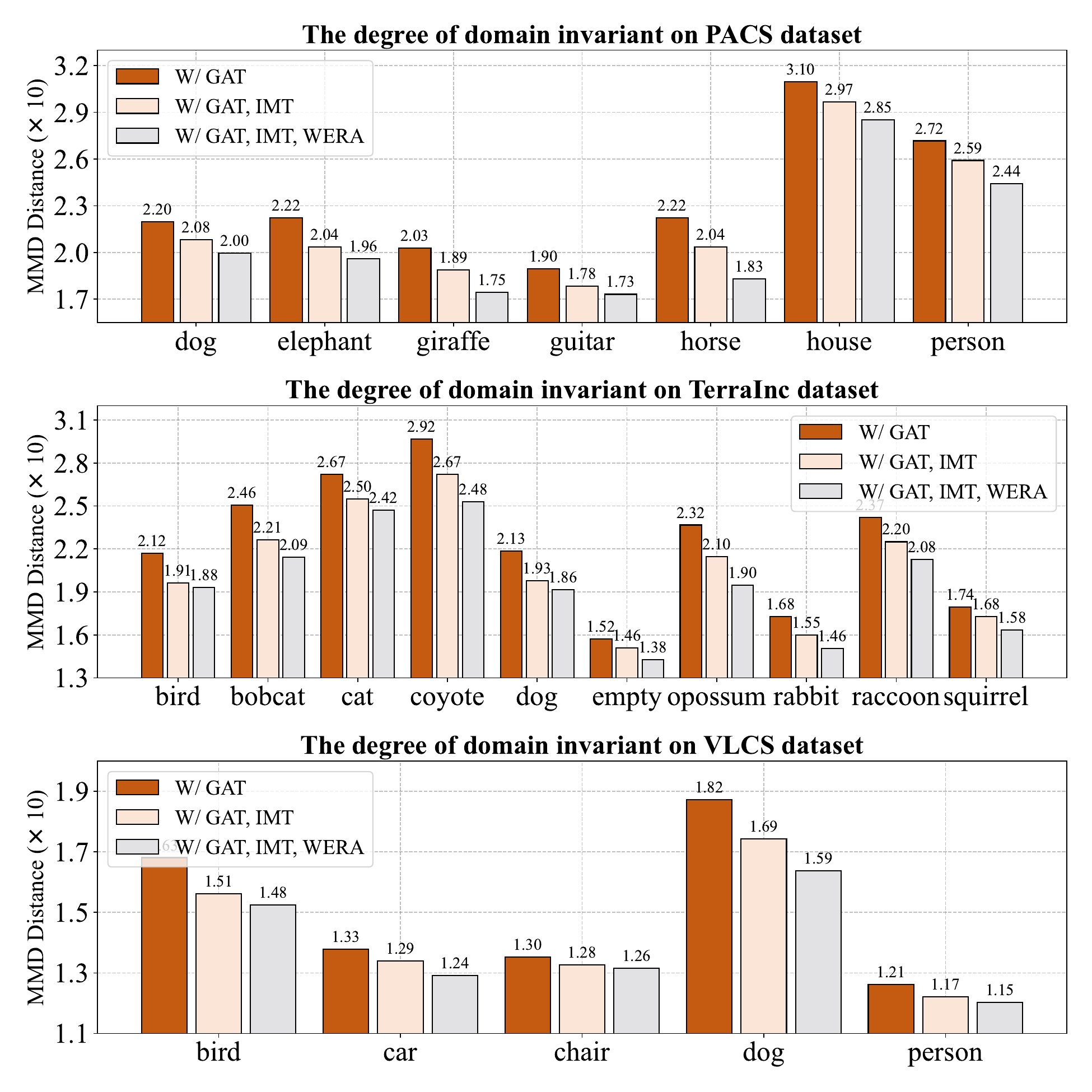}
    \vspace{-4.0mm}
    \caption{
    Comparisons of average source-target divergences ($\times 10$) between the baseline only with the GAT module (W/ GAT), the proposed IMT module (W/ IMT), and the proposed WERA module (W/ WERA) three conditions on the PACS, TerraInc, and VLCS datasets.
    }
    \label{pic:invariant_representation}
    \vspace{-3.5mm}
\end{figure}

\textbf{The effectiveness of Image-Disentanglement Guided by Text (IMT).} 
Our IMT sub-module achieves improvements of 0.70\%, 4.47\%, 1.64\%, 8.59\%, and 0.84\% over the baseline (see $1^{st}$ and $3^{rd}$ rows in Tab.~\ref{tab:2_analysis_abolution}), with corresponding two-tailed $p$-values (0.015, $<$0.0001, 0.006, $<$0.0001, 0.002), all below 0.05, confirming that these gains are significant across all datasets.
When combined with the GAT sub-module, IMT achieves further improvements of 0.60\%, 2.98\%, 0.48\%, 8.84\%, and 0.34\% (see $2^{nd}$ and $7^{th}$ rows in Tab.~\ref{tab:2_analysis_abolution}), with $t$-values ranging from 5.47 to 36.50, also indicating high statistical significance ($p < 0.01$).
These results validate that IMT effectively reduces the distributional discrepancies between different domains and yields more consistent, semantically disentangled visual representations under textual guidance.
To evaluate the degree of domain invariance between each source domain and the target domain, we here measured the estimated maximum mean discrepancy (MMD~\cite{staib2019distributionally}) among source and target domains for each category on PACS, VLCS, officeHome, and TerraInc datasets.
Specifically, let $\{\mathbf{x}_i,y_i\}_{i=1}^{N_m^s}$ and $\{\mathbf{x}^{'}_i,y^{'}_i\}_{i=1}^{N^t}$ be two samples of size $N_m^s$ and $N^t$ drawn i.i.d from $m_{th}$ source domain $\hat{\mathcal{D}}^S_m$ and target domain $\hat{\mathcal{D}}^{\mathcal{T}}$, respectively. 
The MMD distance between pairwise distribution $\hat{\mathcal{D}}^S_m$ and $\hat{\mathcal{D}}^{\mathcal{T}}$ can be formulated as:
\begin{equation}
\begin{aligned}
    \hat{d}^2(\mathcal{D}_m^S,&\mathcal{D}^{\mathcal{T}})
    =\frac{1}{(N_m^s)^2}\sum\limits_{i=1}^{N_m^s}\sum\limits_{j=1}^{N_m^s}\|\mathbf{f}_I(\mathbf{x}_i;\mathbf{E}_I)-\mathbf{f}_I(\mathbf{x}_j;\mathbf{E}_I)\|_{\mathcal{H}_{k}} \\
    &+\frac{1}{N_m^s\times N^t}\sum\limits_{i=1}^{N_m^s}\sum\limits_{j=1}^{N^t}\|\mathbf{f}_I(\mathbf{x}_i;\mathbf{E}_I)-\mathbf{f}_I(\mathbf{x}^{'}_j;\mathbf{E}_I)\|_{\mathcal{H}_{k}} \\
    &+\frac{1}{(N^t)^2}\sum\limits_{i=1}^{N^t}\sum\limits_{j=1}^{N^t}\|\mathbf{f}_I(\mathbf{x}^{'}_i;\mathbf{E}_I)-\mathbf{f}_I(\mathbf{x}^{'}_j;\mathbf{E}_I)\|_{\mathcal{H}_{k}}, 
\end{aligned} 
\end{equation}
where $\mathcal{H}_k$ is a RKHS equipped with kernel $k$. 
Then we estimate the average source-target divergence, defined as:
\begin{equation}
    \hat{\gamma}_{MMD}=\frac{1}{N_s}\sum\limits_{m=1}^{N_s}\hat{d}(\hat{\mathcal{D}}_m^S,\mathcal{D}^{\mathcal{T}}).
\end{equation}
The results are shown in Fig.~\ref{pic:invariant_representation_officehome} and Fig.~\ref{pic:invariant_representation}.
In contrast, the lower source-target domain divergence achieved by our proposed IMT proves its effectiveness in mitigating the distributional discrepancy in visual representation. 

\begin{table}[t]
\caption{
A comparison of fixed and learnable stylization prompt $\hat{\mathcal{A}}$ applied to WERA in Eq.~\ref{function:statistics_calculate} across all benchmark datasets. 
All experiments are based on the $7^{th}$ line of Tab.~\ref{tab:2_analysis_abolution}. 
}
\renewcommand\arraystretch{1.1}
\centering
\scalebox{1.0}{
\footnotesize
\begin{tabular}{c|l|ccc}
\hline
\textbf{Idx}
& \textbf{Dataset} 
& \textbf{Fixed $\hat{\mathcal{A}}$} 
& \textbf{Learnable $\mathbf{\gamma}/\mathbf{\beta}$}
& \textbf{Learnable $\hat{\mathcal{A}}$} \\
\hline
1 & PACS 
& 97.20\tsb{\(\pm\)0.12} 
& 97.17\tsb{\(\pm\)0.09}
& 97.57\tsb{\(\pm\)0.11} \\
2 & VLCS
& 85.88\tsb{\(\pm\)0.17}
& 85.95\tsb{\(\pm\)0.16}
& 86.32\tsb{\(\pm\)0.21} \\
3 & OfficeHome 
& 86.23\tsb{\(\pm\)0.09} 
& 86.20\tsb{\(\pm\)0.15}
& 86.77\tsb{\(\pm\)0.11} \\
4 & TerraInc
& 57.10\tsb{\(\pm\)0.38} 
& 57.43\tsb{\(\pm\)0.23} 
& 58.35\tsb{\(\pm\)0.20} \\
5 & DomainNet
& 62.01\tsb{\(\pm\)0.13} 
& 62.05\tsb{\(\pm\)0.19}
& 62.53\tsb{\(\pm\)0.15} \\
\hline
6 & Avg.
& 77.68\tsb{\(\pm\)0.18}
& 77.76\tsb{\(\pm\)0.17}
& 78.28\tsb{\(\pm\)0.17} \\
\hline
\end{tabular}
}
\label{tab5:analysis_rrl}
\vspace{-2.5mm}
\end{table}
\begin{table}[t]
\caption{
A comparison of training and training-free conditions applied to DSPL across the PACS, VLCS, OfficeHome, TerraInc, and DomainNet datasets. DSPL$^*$ represents the training-free condition applied to DSPL.
All experiments are based on the $8^{th}$ line of Tab.~\ref{tab:2_analysis_abolution}.
$\Delta$: the differences between two conditions.
}
\renewcommand\arraystretch{1.1}
\centering
\scalebox{1.0}{
\footnotesize
\begin{tabular}{c!{\vrule width0.5pt}l!{\vrule width0.5pt}@{\hspace{18pt}}c@{\hspace{18pt}}c@{\hspace{18pt}}!{\vrule width0.5pt}>{\columncolor{gray!30}}c}
\hline
\textbf{Idx}
& \textbf{Dataset} 
& \textbf{DSPL$^{*}$} 
& \textbf{DSPL}
& \textbf{$\Delta$} \\
\hline
1 & PACS 
& 97.65\tsb{\(\pm\)0.15} 
& 97.80\tsb{\(\pm\)0.13} 
& \textcolor{red}{+0.15} \\
2 & VLCS
& 86.44\tsb{\(\pm\)0.20}
& 86.72\tsb{\(\pm\)0.21} 
& \textcolor{red}{+0.28} \\
3 & OfficeHome 
& 86.91\tsb{\(\pm\)0.12} 
& 87.06\tsb{\(\pm\)0.18} 
& \textcolor{red}{+0.15} \\
4 & TerraInc
& 58.62\tsb{\(\pm\)0.23} 
& 59.32\tsb{\(\pm\)0.38} 
& \textcolor{red}{+0.70} \\
5 & DomainNet
& 62.58\tsb{\(\pm\)0.19} 
& 62.74\tsb{\(\pm\)0.14} 
& \textcolor{red}{+0.16} \\
\hline
6 & Avg.
& 78.44\tsb{\(\pm\)0.17}
& 78.73\tsb{\(\pm\)0.21}
& \textcolor{red}{+0.29} \\
\hline
\end{tabular}
}
\label{tab5:analysis_DSPL}
\end{table}

\textbf{The effectiveness of Worst Explicit Representation Alignment (WERA).}
Our WERA module achieves improvements of 1.28\%, 4.93\%, 2.11\%, 10.24\%, and 1.68\% over the baseline (see $1^{st}$ and $4^{th}$ rows in Tab.~\ref{tab:2_analysis_abolution}), with corresponding $t$-values ranging from 9.17 to 52.90 and all two-tailed $p$-values below 0.01, confirming statistically significant gains across all datasets.
When combined with the GAT sub-module, WERA further improves performance by 1.05\%, 3.63\%, 1.16\%, 10.37\%, and 1.03\% (see $2^{nd}$ and $8^{th}$ rows in Tab.~\ref{tab:2_analysis_abolution}), with $t$-values between 11.66 and 35.50 and all $p$-values $<$ 0.01, demonstrating strong statistical significance.
This observation indicates that we can better mitigate the distribution discrepancies between the source and target domain.
Furthermore, we conduct additional experiments to evaluate whether the proposed stylization prompt $\hat{\mathcal{A}}$ can provide further improvements compared with the existing data augmentation method (i.e., MixStyle).
As shown in Tab.~\ref{tab5:analysis_rrl}, incorporating the learnable stylization prompt $\hat{\mathcal{A}}$ leads to a consistent performance gain, surpassing both the learnable $\gamma/\beta$ formulation and the fixed $\hat{\mathcal{A}}$ variant by 0.52\% and 0.60\% in average accuracy across all datasets, respectively.
Additionally, the visualization of source-target divergence in Fig.~\ref{pic:invariant_representation} further validate the WERA module's efficacy in aligning domain distributions.

\begin{figure}[t]
    \centering
    \includegraphics[width=0.49\textwidth]{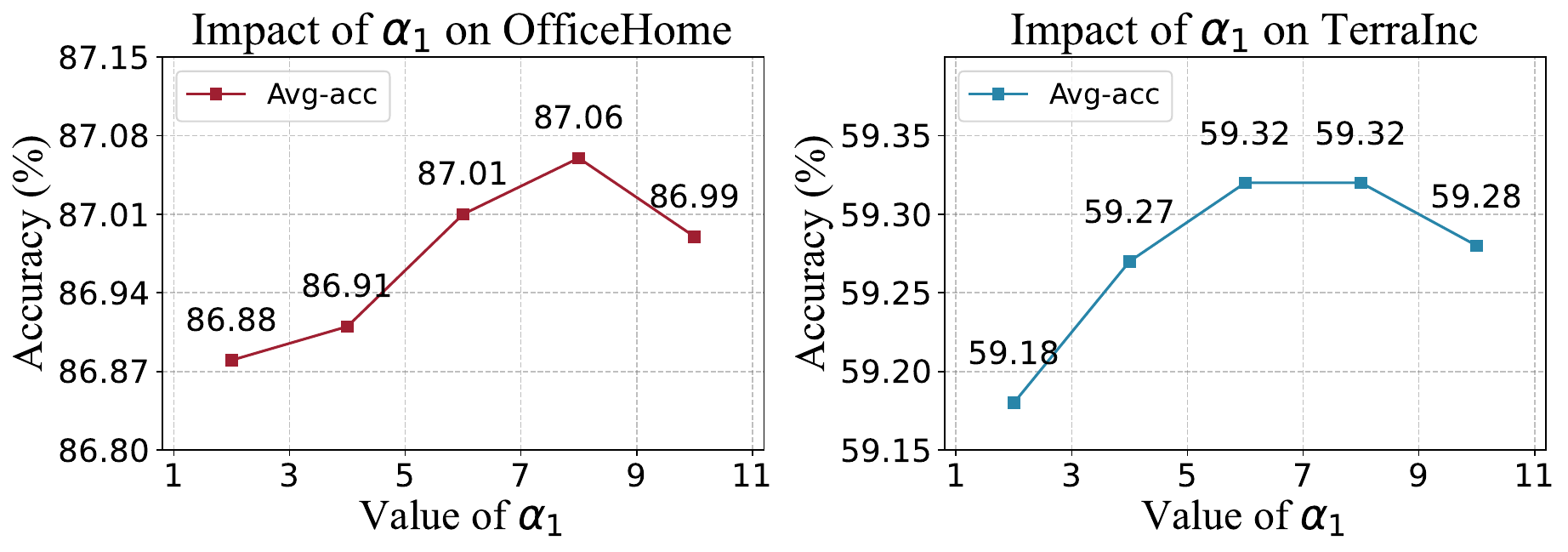}
    \caption{
    Analysis of $\alpha_1$ in Eq.~\ref{loss_1} on OfficeHome and TerraInc datasets for text disentanglement (GAT sub-module).
    }
    \label{pic:parameter_analysis_alpha1}
\end{figure} 

\begin{figure}[t]
    \centering
    \includegraphics[width=0.49\textwidth]{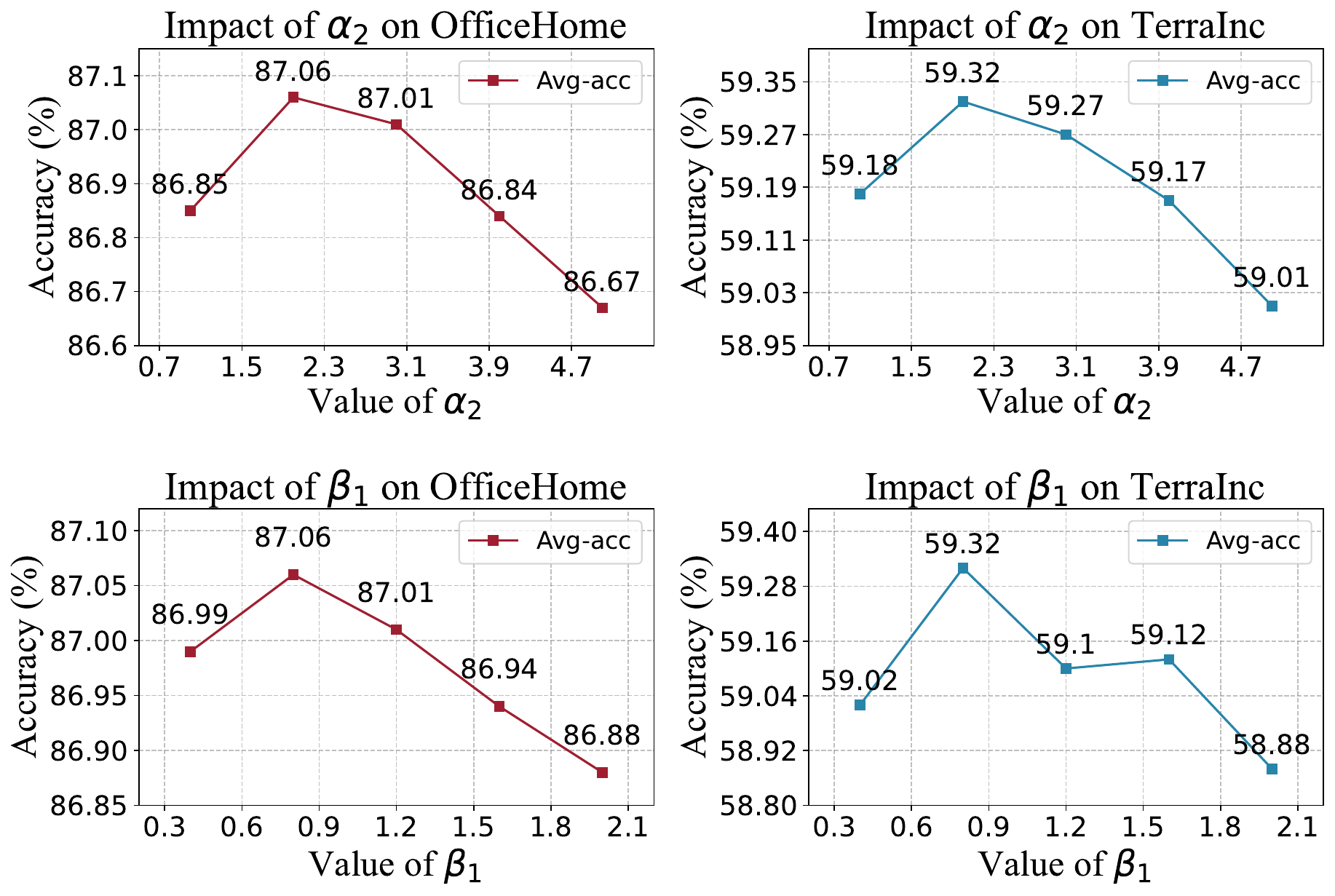}
    \caption{
    Analysis of $\alpha_2$ and $\beta_1$ in Eq.~\ref{loss_2} on OfficeHome and TerraInc datasets for image disentanglement (IMT sub-module).
    }
    \label{pic:parameter_analysis_alpha2}
\end{figure}

\begin{figure}[!htbp]
    \centering
    \includegraphics[width=0.49\textwidth]{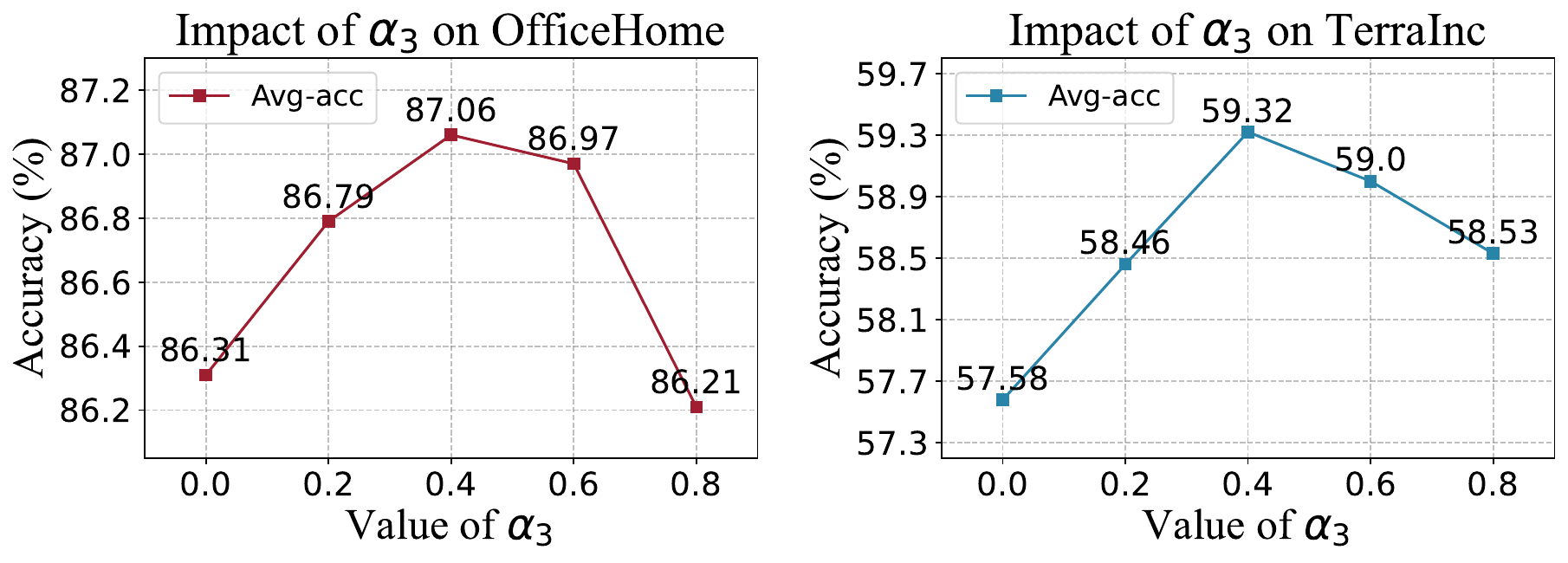}
    \caption{
    Analysis of $\alpha_3$ in Eq.~\ref{function:loss_rrl} on OfficeHome and TerraInc datasets for Worst Explicit Representation Alignment (WERA module).
    }
    \label{pic:parameter_analysis_alpha3}
\end{figure}

\begin{figure}[!htbp]
    \centering
    \includegraphics[width=0.485\textwidth]{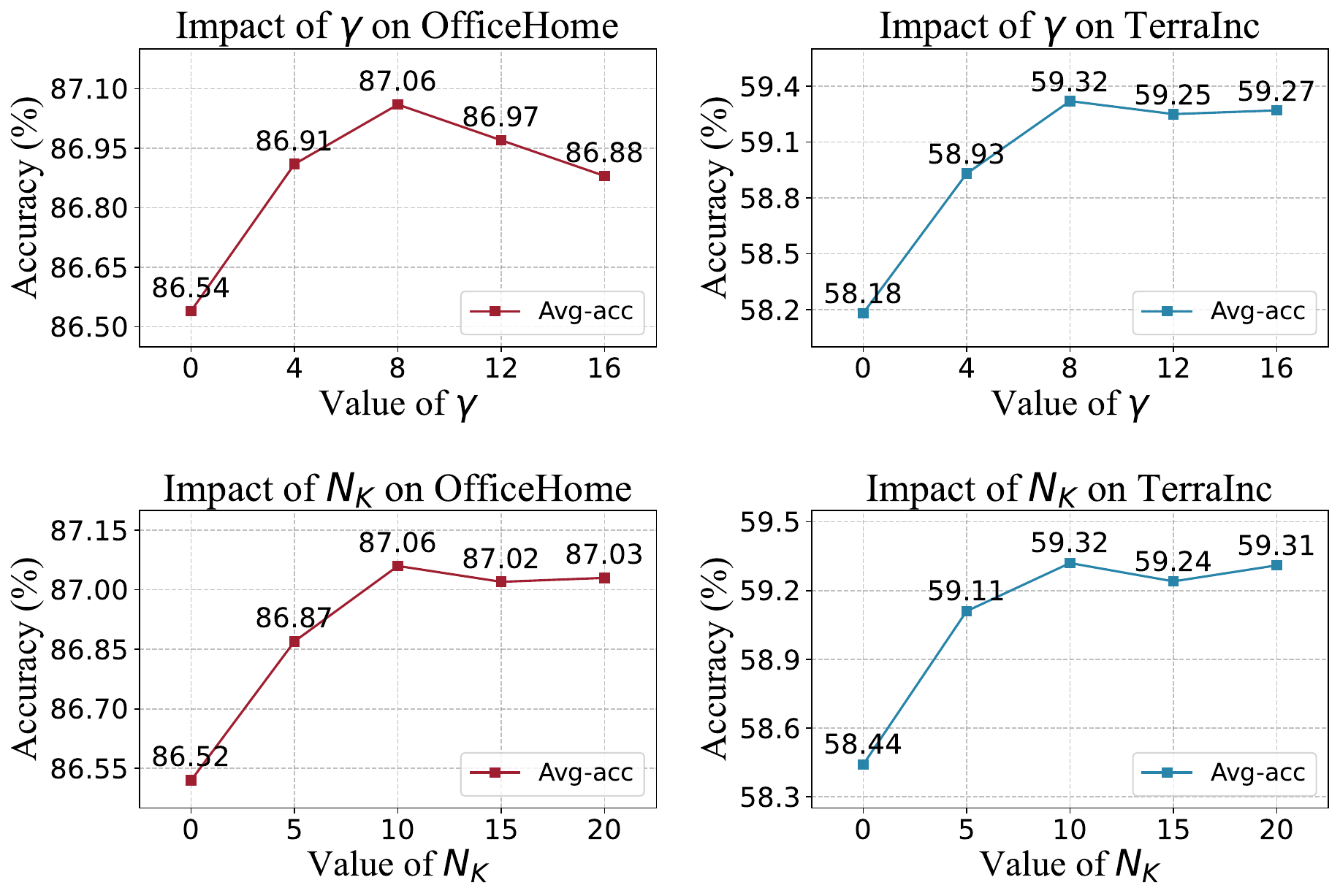}
    \caption{
    Analysis of $\gamma^{'}$ and $N_K$ in Eq.~\ref{function:worst_all} on OfficeHome and TerraInc datasets for Worst Explicit Representation Alignment (WERA module).
    }
    \label{pic:parameter_analysis_gamma_N}
\end{figure}

\begin{figure}[!htbp]
    \centering
    \includegraphics[width=0.49\textwidth]{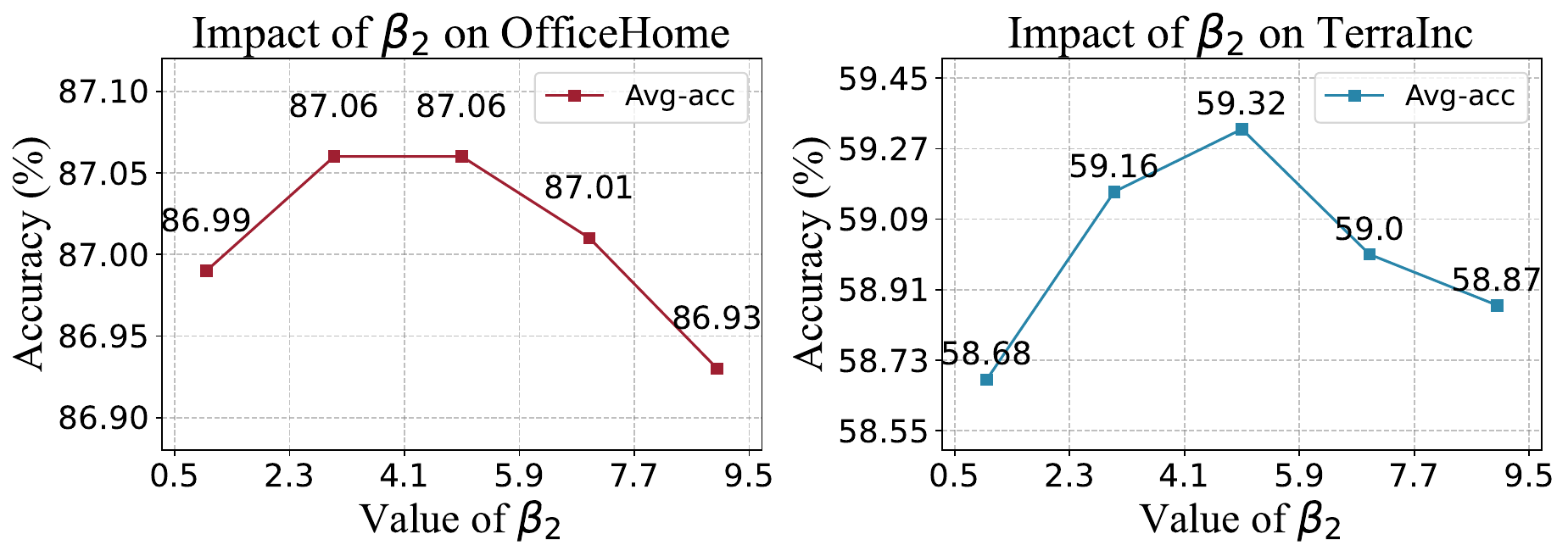}
    \vspace{-3.5mm}
    \caption{
    Analysis of $\beta_2$ in Eq.~\ref{DS_Prediction} on OfficeHome and TerraInc datasets for Domain Specific Prototype Learning (DSPL module).
    }
    \label{pic:parameter_analysis_beta2}
    \vspace{-3.5mm}
\end{figure}

\begin{figure}[t]
\centering
\includegraphics[width=0.48\textwidth]{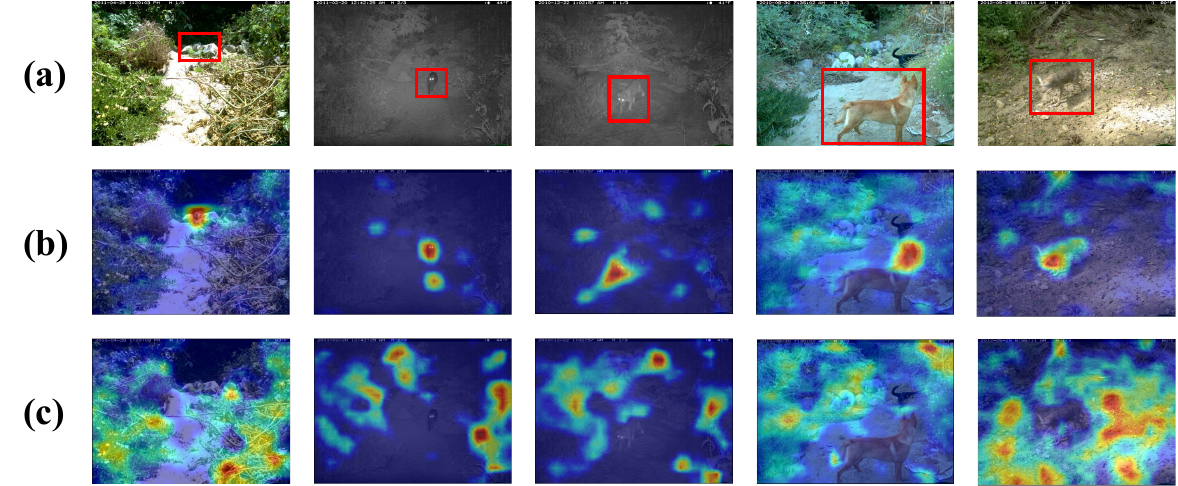}
\caption{
Visualization of attention maps between domain-invariant visual features (\textcolor{red}{b}) and domain-specific visual features (\textcolor{red}{c}) on Terrainc dataset.
}
\label{fig:visualize_pacs}
\end{figure}

\begin{figure}[t]
\centering
\includegraphics[width=0.48\textwidth]{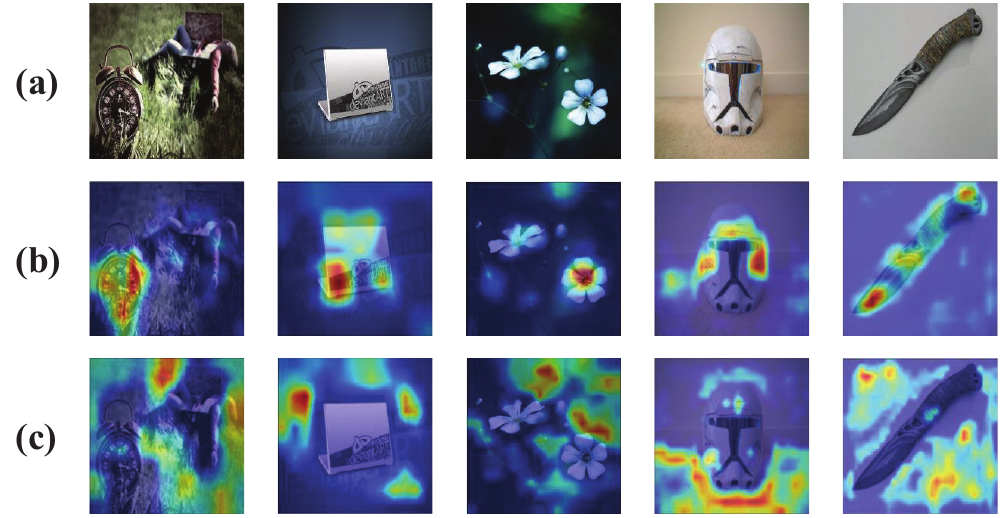}
\caption{
Visualization of attention maps between domain-invariant visual features (\textcolor{red}{b}) and domain-specific visual features (\textcolor{red}{c}) on OfficeHome dataset.
}
\label{fig:visualize_officehome}
\vspace{-3.5mm}
\end{figure}

\textbf{The effectiveness of Domain-Specific Prototype Learning (DSPL).}
The DSPL module provides a performance gain of 0.34\%, 0.82\%, 0.19\%, 0.65\% and 0.47\% when directly adding it to our baseline (see $3^{rd}$ row and $5^{th}$ row in Tab.~\ref{tab:2_analysis_abolution}). When based on the GAT sub-module, the DSPL further improves the performance of 0.36\%, 0.93\%, 0.22\%, 0.68\% and 0.21\% (see $7^{th}$ row and $9^{th}$ row in Tab.~\ref{tab:2_analysis_abolution}). 
This observation indicates that utilizing partial domain-specific knowledge as supplementary information is advantageous for classification. 
Unlike existing disentanglement-based methods, our DSPL framework effectively integrates both domain-invariant and domain-specific information for final inference, leading to a demonstrable improvement in generalization performance.
Furthermore, we conduct experiments to evaluate the effectiveness of DSPL under two different conditions: `training-free' and `training' in Tab.~\ref{tab5:analysis_DSPL}. 
The results show that, under the given training conditions, the performance improvement achieved by DSPL is 0.15\%, 0.28\%, 0.15\%, 0.70\%, and 0.16\%, respectively, compared to the training-free baseline.

\subsection{Hyper-Parameter Analysis}
We conduct experiments to analyze the impact of the hyper-parameters in our method on OfficeHome and TerraInc benchmark datasets, including $\alpha_1$ in Eq.~\ref{loss_1}, $\alpha_2$, $\beta_1$ in Eq.~\ref{loss_2}, $\gamma^{'}$ in Eq.~\ref{function:worst_all}, $\alpha_3$ Eq.~\ref{function:loss_rrl}, $N_K$ in Sec.~\ref{worst-case}, $\beta_2$ in Eq.~\ref{prototype_weight}.

(1) We first analyze the hyper-parameters in GAT, including $\alpha_1$. The results are shown in Fig.~\ref{pic:parameter_analysis_alpha1}.
The parameter $\alpha_1$ serves to balance the importance of the regularization term during text disentanglement.
The higher $\alpha_1$ denotes that the domain-invariant textual embeddings pay more attention to the general knowledge generated by handcraft descriptions.
Based on the results, we select $\alpha_1=8.0$ for all benchmark datasets. 
(2) Then we analyze the hyper-parameter in IMT, including $\alpha_2$, and $\beta_1$.
The results are shown in Fig.~\ref{pic:parameter_analysis_alpha2}.
The parameter $\alpha_2$ serves to balance the importance of the regularization term during the image disentanglement and the parameter $\beta_1$ is the intensity of the domain confusion.
Based on the results, we select $\alpha_2=2.0$ and $\beta_1=0.8$ for all benchmark datasets.
(3) Next, we analyze the hyper-parameteru in WERA, including $\gamma^{'}$, $\alpha_3$, and $N_K$.
The results are shown in Fig.~\ref{pic:parameter_analysis_alpha3} and Fig.~\ref{pic:parameter_analysis_gamma_N}.
Based on the results, we select $\gamma^{'}=8.0$, $\alpha_3=0.4$, and $N_K=10$ for all benchmark datasets.
$\alpha_3$ serves to balance the loss term on the original distribution and worst-case distribution. 
(4) Finally, we analyze the hyper-parameter in DSPL, including $\beta_2$. 
The results are shown in Fig.~\ref{pic:parameter_analysis_beta2}.
Based on the results, we select $\beta_2=5.0$ for all benchmark datasets.

\begin{table*}[t]
\small
\centering
\caption{
\footnotesize{
Resource consumption comparison on the TerraInc dataset.
\textit{Memory usage (GB)} denotes the total GPU memory occupied during training, with a batch size of 16. 
\textit{Training Params (MB)} denotes the number of parameters of each method.
\textit{Training time (ms)} and \textit{Inference time (ms)} are measured per sample.
\textit{Inference GFLOPs} refers to the number of floating point operations required for inference per sample.
All experiments are conducted on a single NVIDIA RTX 3090 GPU.
}}
\label{tab:cost}
\scalebox{0.85}{
\begin{tabular}{l|ccc|ccc}
\toprule
\multirow{2}{*}{\textbf{Method}} & 
\multicolumn{3}{c|}{\textbf{Trainable stage}} & 
\multicolumn{3}{c}{\textbf{Inference stage}} \\
\cmidrule(lr){2-4} \cmidrule(lr){5-7}
& \textbf{Memory usage (GB)}
& \textbf{Training Params (MB)} & \textbf{Training time (ms/img)} 
& \textbf{Inference GFLOPs}
& \textbf{Inference time (ms/img)} & \textbf{Acc. (\%)} \\
\midrule
CoOp~\cite{Zhou_2022}
& 1.958 & 0.016 & 3.22 & 94.87 & 1.40 & 48.80 \\ 
Maple~\cite{khattak2023maple}
& 2.330 & 3.68 & 3.71 & 108.25 & 1.65 & 50.20 \\
\midrule
DPR~\cite{cheng2024disentangled}   
& 2.016 & 0.15 & 3.57 & 101.55 & 1.53 & 57.10 \\
PADG (ours)      
& 2.452 & 0.34 & 4.34 & 101.55 & 1.53 & 59.32 \\
\bottomrule
\end{tabular}}
\end{table*}

\subsection{Further Analysis}
\noindent \textbf{Visualization of the image disentanglement.}
To demonstrate the effectiveness of our method in mitigating domain discrepancy through image disentanglement, we visualize the feature maps using the Grad-Cam~\cite{selvaraju2017grad} algorithm, as shown in Fig.\ref{fig:visualize_pacs} and Fig.\ref{fig:visualize_officehome}. 
The domain-invariant features include areas like the eye, mouth, mustache, etc, while the domain-specific features include the background, etc.
The results reveal that our method successfully disentangles the visual representation, enabling the model to focus on domain-invariant and domain-specific features respectively, further validating the efficacy of our image disentanglement (IMT sub-module).

\noindent \textbf{Computational cost.}
We further analyze the computational and resource overhead introduced by the WERA module by comparing the proposed PADG with the DPR version from CVPR’24 in terms of runtime and resource usage (see Tab.~\ref{tab:cost}). All results were conducted on the TerraInc dataset using a single NVIDIA RTX 3090 GPU (24 GB memory).
As shown in Tab.~\ref{tab:cost}, adding WERA increases the GPU memory usage by 0.46 GB (from 2.016 GB in DPR to 2.452 GB in PADG), the trainable parameters by only 0.19 MB (from 0.15 MB in DPR to 0.34 MB in PADG), and training time per sample by 0.77 ms (a 21.6\% relative increase), while inference time remains unchanged at 1.53 ms.
Compared to other methods, CoOp~\cite{Zhou_2022} has the smallest parameters (0.016 MB) but achieves the lowest accuracy (48.80\%), while Maple~\cite{khattak2023maple} has a significantly larger parameter size (3.68 MB) and a longer training time (3.71 ms), with an accuracy of 50.20\%. In contrast, PADG demonstrates a +2.22\% accuracy improvement over DPR (which has an accuracy of 57.10\%) with minimal overhead in parameters and training time. These results highlight that WERA provides robust domain generalization with negligible computational cost.
Overall, the PADG framework maintains high computational efficiency while significantly improving both robustness and generalization capability.

\section{Conclusion}
In this paper, we reframe the prompt tuning framework by adopting a disentanglement perspective for domain generalization. 
Specifically, we first introduce the Cross-Modal Prompt Disentanglement module, which learns disentangled domain-invariant and domain-specific visual features guided by disentangled textual features. 
Subsequently, we propose the Worst Explicit Representation Alignment module to further mitigate the domain discrepancies between source and target domains.
To fully leverage both the domain-invariant information and domain-specific information for the final inference, we propose a novel ensemble learning strategy: Domain-Specific Prototype Learning module. 
Extensive experiments on the DomainBed dataset have demonstrated the effectiveness of our proposed method and each component. 

Despite these promising results, our current WERA module perturbs features mainly at the statistical level (e.g., mean and variance), which may not fully capture complex or non-linear domain shifts. In future work, we plan to explore more fine-grained and semantically aware perturbation strategies that model higher-order feature dependencies. Moreover, we aim to extend our framework toward a more robust cross-modality prompt disentanglement paradigm, which remains a fundamental challenge for domain generalization.
\bibliographystyle{IEEEtran}
\bibliography{egbib}

\begin{IEEEbiography}[{\includegraphics[width=1in,height=1.25in,clip,keepaspectratio]{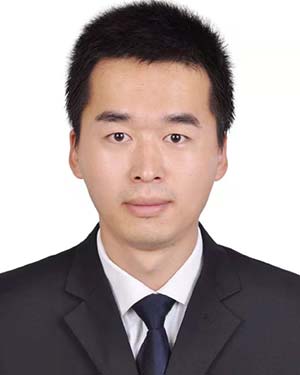}}]{De Cheng} is an associate professor with School of Telecommunications Engineering, Xidian University, China. He received the B.S. and Ph.D. degrees from Xi'an Jiaotong University, Xi'an, China, in 2011 and 2017, respectively. From 2015 to 2017, he was a visiting scholar in Carnegie Mellon University, Pittsburgh, USA. His research interests include pattern recognition, machine learning, and multimedia analysis.
\end{IEEEbiography}

\begin{IEEEbiography}[{\includegraphics[width=1in,height=1.25in,clip,keepaspectratio]{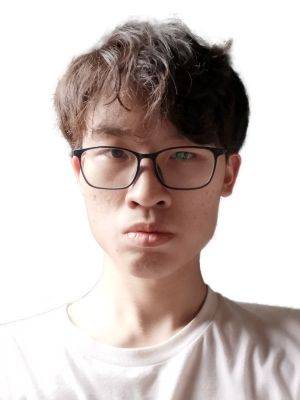}}]
{Zhipeng Xu} received the B.Sc. degree from Xidian University, Xi'an, China, in 2023. He is currently pursuing his M.S. degree in Information and Communication Engineering in Xidian University. His research interests in domain generalization, few-shot learning, and prompt tuning.
\end{IEEEbiography}

\begin{IEEEbiography}[{\includegraphics[width=1in,height=1.25in,clip,keepaspectratio]{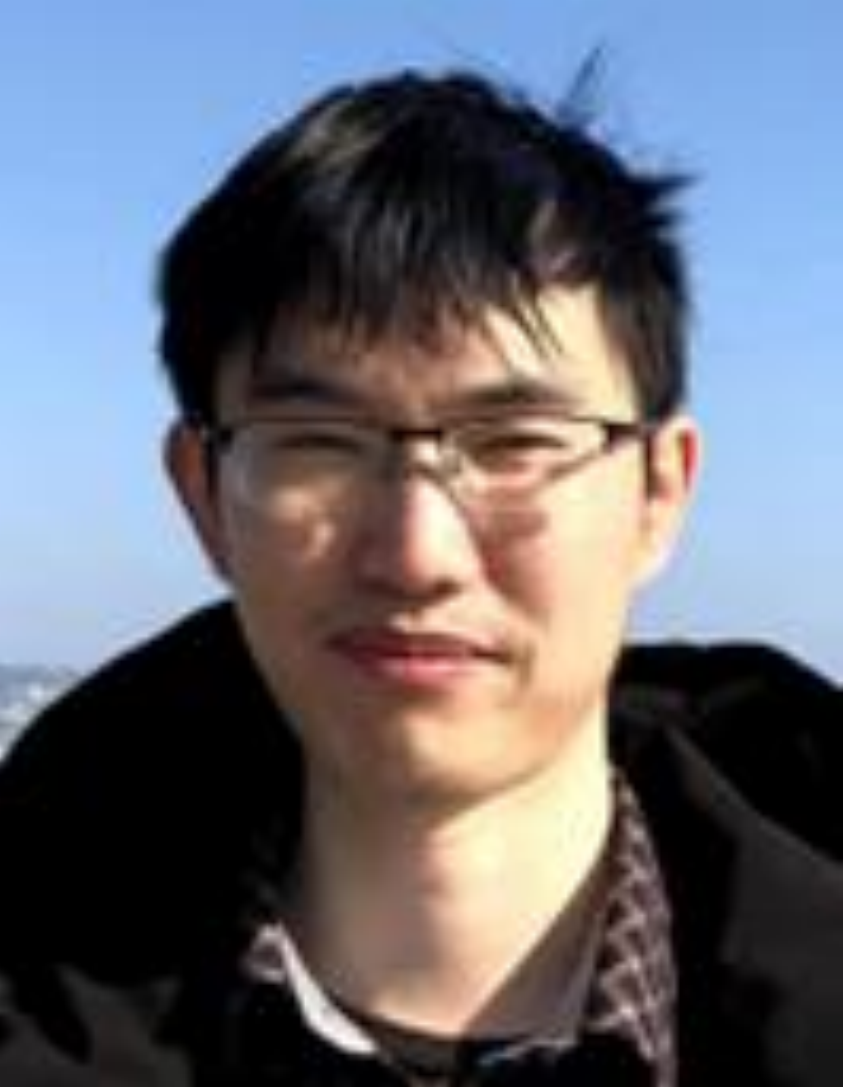}}]{Xinyang Jiang} received B.E. from Zhejiang University in 2012 and Ph.D. from Zhejiang University in 2017. He is now a researcher from Microsoft Research Asia. Before joining MSRA, he was a researcher from Tencent Youtu Lab. His main research field is computer vision, including person Re-identification, vector graphics recognition and medical image understanding.
\end{IEEEbiography}

\begin{IEEEbiography}[{\includegraphics[width=1in,height=1.25in,clip,keepaspectratio]{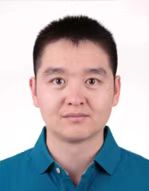}}]{Dongsheng Li} received B.E. from University of Science and Technology of China in 2007 and Ph.D. from Fudan University in 2012. He is now a principal research manager with Microsoft Research Asia (MSRA) since February 2020.  Before joining MSRA, he was a research staff member with IBM Research – China  since April 2015. He is also an adjunct professor with  School of Computer Science, Fudan University, Shanghai, China. His research interests include recommender systems and machine learning applications. His work on cognitive recommendation engine won the 2018 IBM Corporate Award.
\end{IEEEbiography}

\begin{IEEEbiography}[{\includegraphics[width=1in,height=1.25in,clip,keepaspectratio]{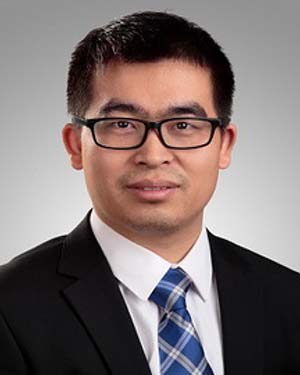}}]{Nannan Wang}(M'16) received the B.Sc. degree in information and computation science from the Xi¡¯an University of Posts and Telecommunications in 2009 and the Ph.D. degree in information and telecommunications engineering from Xidian University in 2015. From September 2011 to September 2013, he was a Visiting Ph.D. Student with the University of Technology, Sydney, NSW, Australia. He is currently a Professor with the State Key Laboratory of Integrated Services Networks, Xidian University. He has published over 100  articles in refereed journals and proceedings, including IEEE T-PAMI, IJCV, CVPR, ICCV etc. His current research interests include computer vision and machine learning.
\end{IEEEbiography}

\begin{IEEEbiography}[{\includegraphics[width=1in,height=1.25in,clip,keepaspectratio]{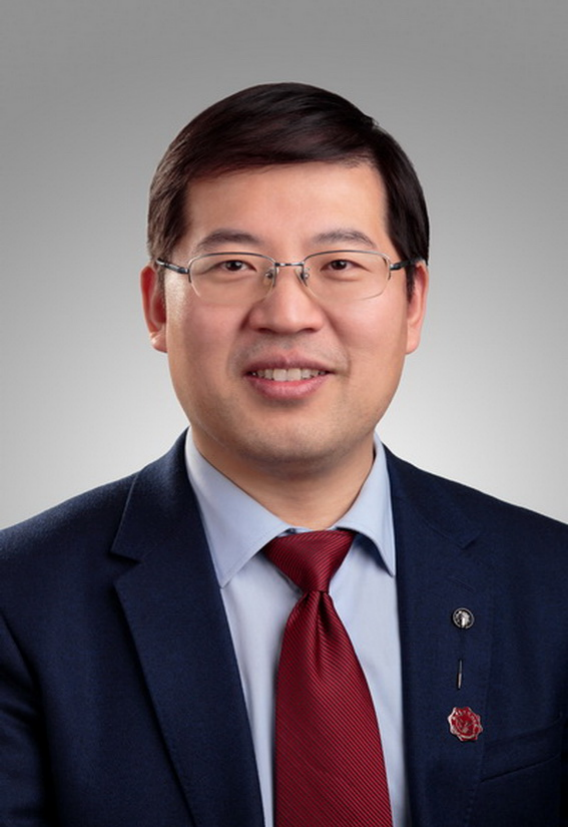}}]{Xinbo Gao}(M'02-SM'07-F'24) earned his B.Eng. degree in Electronic Engineering (1994), followed by M.Sc. (1997) and Ph.D. (1999) degrees in Signal and Information Processing, all from Xidian University, Xi'an, China. From 1997 to 1998, he was a research fellow in the Department of Computer Science at Shizuoka University, Shizuoka, Japan. During 2000-2001, he served as a post-doctoral research fellow in the Department of Information Engineering at the Chinese University of Hong Kong, Hong Kong. Since 1999, he has been affiliated with Xidian University's School of Electronic Engineering, where he currently holds the position of Full Professor in Pattern Recognition and Intelligent Systems. During 2020-2025, he was a Professor of Computer Science and Technology at Chongqing University of Posts and Telecommunications, Chongqing, China. His research focuses on computer vision, machine learning, and pattern recognition. With over 500 peer-reviewed publications and seven academic books to his credit, Prof. Gao serves on the editorial boards of several prestigious journals including Research, Signal Processing (Elsevier), Neurocomputing (Elsevier) etc. He has chaired or participated in program committees for approximately 50 international conferences. Prof. Gao is a Fellow of multiple professional societies: IEEE, IET, WIA, AAIA, CIE, CCF, CAAI and CSIG.
\end{IEEEbiography}

\end{document}